\documentclass[10pt,twocolumn,letterpaper]{article}

\usepackage[pagenumbers]{wacv} % To force page numbers, e.g. for an arXiv version
\usepackage[accsupp]{axessibility}

\usepackage{graphicx}
\usepackage{amsmath}
\usepackage{amssymb}
\usepackage{booktabs}
\usepackage{multirow}
\usepackage{array,cite}
\usepackage{enumitem}
\usepackage{xcolor}
\usepackage{colortbl} 
\usepackage{makecell}
\usepackage{ragged2e}
\usepackage{placeins}
\usepackage{float}

\usepackage[pagebackref,breaklinks,colorlinks]{hyperref}

\usepackage[capitalize]{cleveref}
\crefname{section}{Sec.}{Secs.}
\Crefname{section}{Section}{Sections}
\Crefname{table}{Table}{Tables}
\crefname{table}{Tab.}{Tabs.}

\def\wacvPaperID{1233}
\def\confName{WACV}
\def\confYear{2025}

\begin{document}

%%%%%%%%% TITLE 
\title{Generative Model-Based Fusion for Improved Few-Shot Semantic Segmentation of Infrared Images}

\author{Junno Yun, Mehmet Ak{\c{c}}akaya\\
University of Minnesota\\
{\tt\small $\{$yun00049, akcakaya$\}$@umn.edu}
}
\maketitle

%%%%%%%%%% ABSTRACT 
\begin{abstract}
  Infrared (IR) imaging is commonly used in various scenarios, including autonomous driving, fire safety and defense applications. Thus, semantic segmentation of such images is of great interest. However, this task faces several challenges, including data scarcity, differing contrast and input channel number compared to natural images, and emergence of classes not represented in databases in certain scenarios, such as defense applications. Few-shot segmentation (FSS) provides a framework to overcome these issues by segmenting query images using a few labeled support samples. However, existing FSS models for IR images require paired visible RGB images, which is a major limitation since acquiring such paired data is difficult or impossible in some applications. In this work, we develop new strategies for FSS of IR images by using generative modeling and fusion techniques. To this end, we propose to synthesize auxiliary data to provide additional channel information to complement the limited contrast in the IR images, as well as IR data synthesis for data augmentation. Here, the former helps the FSS model to better capture the relationship between the support and query sets, while the latter addresses the issue of data scarcity. Finally, to further improve the former aspect, we propose a novel fusion ensemble module for integrating the two different modalities. Our methods are evaluated on different IR datasets, and improve upon the state-of-the-art (SOTA) FSS models. 
\end{abstract}

 \begin{figure}[tb]
  \centering
  \includegraphics[width=0.46\textwidth]{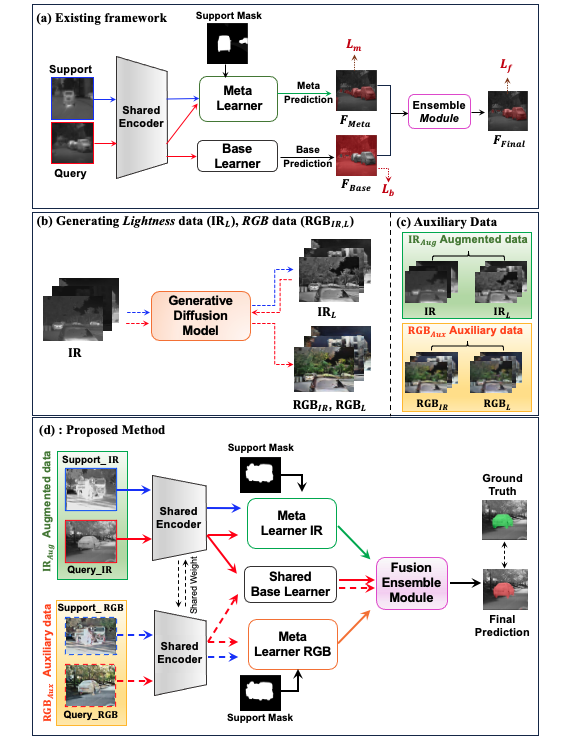}
  \vspace{-0.2cm}
  \caption{Overall architectures of the proposed methods. (a) We utilize the existing FSS model for IR images. (b), (c) To improve this model, we generate new \textit{lightness} and \textit{RGB} datasets with a generative diffusion model.  (d) To exploit them, we propose an additional meta-learning component and fusion networks.}
  \label{fig:overview}
  \vspace{-0.6cm}
\end{figure}

%%%%%%%% Introduction 
\section{Introduction}
\label{sec:intro}
Semantic segmentation classifies each pixel of an image to a specific category, providing precise scene comprehension and object localization at the pixel level across numerous applications~\cite{wang2018understanding}. While RGB images are ubiquitous in computer vision, infrared (IR) images capture temperature-related details, enhancing visibility in challenging conditions such as low illumination, smog, and fog~\cite{danaci2023survey, kutuk2022semantic, li2020segmenting}. Thus, visual/RGB and IR images offer complementary information. However, in numerous applications, acquiring visual and IR data concurrently is impossible, including in defense systems~\cite{d2019cnn}, maritime object recognition~\cite{wang2015infrared, khellal2018convolutional}, and fire safety~\cite{kim2015real, bhattarai2020deep}. This has led to interest in semantic segmentation of standalone IR images~\cite{li2020segmenting, panetta2021ftnet}.

However, IR image segmentation models face several challenges. First, obtaining large-scale annotated IR datasets is difficult due to expensive sensors and restricted access to data from defense and commercial applications~\cite{danaci2023survey}. This scarcity of annotated IR datasets limits the training of deep learning (DL) models, particularly with supervised learning~\cite{bustos2023systematic}. Additionally, unseen or rare classes are common in IR imaging applications, posing a challenge for supervised DL models to generalize to new classes~\cite{shaban2017one, vinyals2016matching}. Lastly, thermal cameras typically produce low-resolution, low-contrast IR images with unclear boundaries~\cite{li2020segmenting}, and the single gray-scale channel provides limited information compared to RGB’s three channels, which leads to performance issues in existing segmentation models~\cite{panetta2021ftnet, xiong2021mcnet, li2020segmenting}.

Few-shot segmentation (FSS) addresses these challenges by learning effective segmentation from a small number of labeled examples, potentially mitigating the lack of training data and improving generalization to unseen classes~\cite{shaban2017one,rakelly2018few,dong2018few,zhang2020sg,wang2019panet,tian2020prior,lang2022learning, vinyals2016matching}. This approach is particularly critical for defense applications of IR, as the increasing complexity of battlefield environments introduces new categories of targets, such as UAVs, weapons, and obstacles. This relevance also extends to firefighting. However, FSS models have found limited use for IR data due to issues including scarcity of IR data and the lower number of channels containing information compared to RGB images~\cite{bao2021visible, zhang2022adfnet, zhao2023bmdenet}.

In cases where real data is insufficient, as is often the case with IR datasets, synthetic data can enhance DL model performance~\cite{yang2022image}. While image-to-image (I2I) translation methods can augment IR images from RGB images, they depend on RGB images and corresponding segmentation masks, requiring further annotation~\cite{yun2019improved}, especially if they lack corresponding masks. Thus, augmenting IR data while maintaining semantic categories poses a challenge. 

On the other hand, to address channel number differences, various fusion networks~\cite{ha2017mfnet, sun2019rtfnet, sun2020fuseseg} have been proposed to leverage IR image properties by fusing them with RGB data. Combining IR and visible spectra can synergistically enhance segmentation models, as they provide distinct insights. Multi-modal FSS frameworks~\cite{bao2021visible, zhang2022adfnet, zhao2023bmdenet} have been developed for paired RGB-IR datasets, improving IR image segmentation. However, these methods are limited to \emph{paired RGB-IR datasets}, which are difficult to obtain in practice.

\looseness=-1
To tackle these challenges, we propose to use generative DL models along with associated novel fusion networks as an alternative strategy for synthesizing images from unpaired data~\cite{goodfellow2020generative, zhu2017unpaired}, as depicted in~\cref{fig:overview}(b-c), aiming to improve the performance of FSS models for IR images without requiring paired RGB data. To this end, we propose the use of two generative DL methods for enhancing the performance. First, we generate \textit{lightness images} (L value from the LAB color space) from existing IR images, maintaining the same gray channel while featuring more distinct boundaries, thereby augmenting the training data and mitigating the scarcity of IR data in public databases without additional annotation work. Second, we synthesize \textit{RGB images} from the IR and lightness images generated in the initial steps and utilize them as auxiliary information through an additional meta-learning component to emphasize color space similarity. Afterward, a novel fusion process combines predictions from the two meta learners and the base learner, as shown in~\cref{fig:overview}(d).

Our FSS approach for IR images is evaluated on SODA (Segmenting Objects in Day And night)\cite{li2020segmenting} and SCUTSEG\cite{xiong2021mcnet} datasets, demonstrating improvement over existing strategies.

Our main contributions are \vspace{-7pt}:
\begin{itemize}[itemsep=0.1em, parsep=0pt]
\item We propose using \textit{generated lightness images} as augmented training data to address the scarcity of IR data. Additionally, we exploit \textit{multi-modal data augmentation} by generating \textit{synthesized RGB images} to provide additional channel information. 
\item We adapt the FSS task for IR images, and propose an FSS model with a novel meta learner and a \emph{multi-modal fusion ensemble} module.
\item Our method achieves significant improvements on SODA and SCUTSEG datasets compared to SOTA FSS models. 
\item We demonstrate that utilizing fusion techniques with generative methods yields promising results \textit{without paired RGB-IR data}.
\end{itemize}

%%%%%%%%%% Related Work   
\section{Related Work}
In this section, we provide background on FSS methods for RGB and IR images, as well as I2I translation with generative diffusion models.

\subsection{Few-Shot Segmentation}  \label{sec21}
FSS is a few-shot learning strategy for segmenting objects of interest in images using only a few annotated samples. A common FSS architecture involves two branches~\cite{shaban2017one}. The first branch generates classifier parameters by processing support images, capturing relevant information. The second branch, the segmentation branch, receives a query image and produces a segmentation mask using the parameters from the first branch. 

Outstanding achievements in FSS have been made by leveraging prototype learning methods, a type of metric-based approach~\cite{koch2015siamese, sung2018learning, snell2017prototypical, vinyals2016matching}. For instance, masked average pooling~\cite{zhang2020sg} enhanced the extraction of class representative prototype vectors from the support set. PFENet~\cite{tian2020prior} introduced a non-parametric prior mask generation method based on cosine similarity of high-level features, improving generalization. HSNet~\cite{min2021hypercorrelation} used hyper-correlation squeeze networks to leverage multi-level feature correlations and employed lightweight 4D convolutions for fine-grained segmentation. An attention-based multi-context guiding network~\cite{hu2019attention} emphasized context information from small to large scales to guide query branches globally. To address biases in previous frameworks reliant on meta-learning, BAM~\cite{lang2022learning} proposed a novel FSS approach with three components: a meta learner designed to recognize novel, unseen classes, following the conventional meta-learning paradigm; a base learner trained in a supervised manner on known base classes; and an ensemble module that integrates coarse predictions from both learners for accuracy.

Building on the architecture of BAM and previous frameworks, MSANet~\cite{iqbal2022msanet} further enhanced the FSS model by introducing two guiding modules into the meta learner: the multi-similarity module and the lightweight attention block. The multi-similarity module calculates visual correspondences between features extracted from intermediate and high-level layers of support and query images, establishing meaningful relationships between the two sets of features. The lightweight attention module, composed of convolutional neural networks, enables the meta learner to focus on the ideal target in the query image.

\subsection{Infrared Image Few-Shot Segmentation}
Segmentation of IR images has seen advances despite its challenges. A major limitation when relying solely on IR images for segmentation has been related to issues such as obscure boundaries, low resolution, and low contrast. To overcome these issues, methods~\cite{li2020segmenting, panetta2021ftnet} applying edge information have been suggested, improving the performance of the segmentation models. Furthermore, visual-thermal spectrum fusion networks~\cite{ha2017mfnet, sun2019rtfnet, sun2020fuseseg, liu2023multi} have been proposed to compensate for the individual limitations of visual and IR images, and to incorporate the strengths of both modalities. As part of these advancements, visual-thermal FSS methods that combine FSS and visual-thermal fusion framework have been proposed to address data sparsity and the growing generalization challenge of unseen classes~\cite{bao2021visible, zhang2022adfnet, zhao2023bmdenet}. However, these approaches still face the problem of acquiring paired IR-RGB data, imposing significant limitations in existing applications. Therefore, an approach that only utilizes IR images for FSS is necessary. 

\subsection{Generative Diffusion Models }
Diffusion models have achieved remarkable success in unsupervised I2I translation~\cite{ho2020denoising, song2020denoising, song2020score, dhariwal2021diffusion}. One such diffusion model, DDPM~\cite{ho2020denoising} generates data by progressively denoising through a series of steps. This model initiates a forward process, which gradually adds Gaussian noise to the data over \(T\) time steps:
\vspace{-0.1cm}
\begin{equation}
    x_t = \sqrt{1 - \beta_t} x_{t-1} + \sqrt{\beta_t} \varepsilon, \quad \varepsilon \sim \mathcal{N}(0, I)
\vspace{-0.05cm}
\end{equation}
\vspace{-0.3cm}
\begin{equation}
    q(x_t | x_{t-1}) = \mathcal{N}(x_t; \sqrt{1 - \beta_t} x_{t-1}, \beta_t I)
\end{equation}
where \(\beta_t\) controls the amount of noise added at each step. Following the forward process, the reverse process gradually generates images from a desired distribution through iterative denoising, maximizing data likelihood:
\vspace{-0.1cm}
\begin{equation}
    p_\theta(x_{t-1} | x_t) = \mathcal{N}(x_{t-1}; \mu_\theta(x_t, t), \Sigma_\theta(x_t, t))
\vspace{-0.1cm}
\end{equation}
where \(\mu_\theta\) is a learned function, and \(\Sigma_\theta\) is known variance with a preset schedule. After training the reverse process, the sampling generates a denoised image \(x_0\) from a noise vector \(x_T\) using reverse diffusion. For each time step \(t\) from \(T\) to \(1\), update the image to approximate \(x_0\) as follows:
\vspace{-0.1cm}
\begin{equation}
    x_{t-1} = \frac{1}{\sqrt{\alpha_t}} \left(x_t-\frac{1-\alpha_t}{\sqrt{1-\bar{\alpha}_t}} \epsilon_\theta(x_t, t) \right) + \sigma_t z_t, z_t \sim \mathcal{N}(0, I)
\label{eq:diff4}
\end{equation}
where \( \alpha_t \) and \( \bar{\alpha}_t \) control the variance schedule, \( \epsilon_\theta(x_t, t) \) is the neural network's noise estimate at step \( t \), and \( \sigma_t \) is the standard deviation of the noise added at each step.

\looseness=-1
Although denoising diffusion models improve sample quality and offer more stable training, they often require numerous inference steps~\cite{alcalar2024ECCV}, presenting computational challenges. 
An alternative approach~\cite{ozbey2023unsupervised} is to utilize
an adversarial conditional diffusion approach, integrating non-diffusive and diffusive models to address these challenges for I2I translation. Here, non-diffusive models utilize generative adversarial networks (GANs) with cycle-consistent architectures for unsupervised training~\cite{goodfellow2014generative, zhu2017unpaired}, generating conditioning images \(\tilde{x}_0\) and \(\tilde{y}_0\) to guide denoising. Meanwhile, diffusive models take these source-conditioning images~\cite{sasaki2021unit} and efficiently sample images using adversarial projectors \(G_x\) and \(G_y\)~\cite{xiao2022tackling} to capture reverse transition probabilities over large step sizes:
\vspace{-0.1cm}
\begin{equation}
    x_t = \sqrt{1 - \gamma_{t}} x_{t-k} + \sqrt{\gamma_{t}} \varepsilon, \quad \varepsilon \sim \mathcal{N}(0, I)
\label{eq:diff5}
\vspace{-0.05cm}
\end{equation}
\begin{equation}
    q(x_t \mid x_{t-k}) = \mathcal{N}(x_t; \sqrt{1 - \gamma_t} x_{t-k}, \gamma_t I)
\vspace{-0.05cm}
\label{eq:diff6}
\end{equation}
where \( k \gg 1 \) denotes the step size, the denoising generators \(G_x\) and \(G_y\) produce \(\hat{x}_0\) and \(\hat{y}_0\) respectively, by implementing the distribution as follows:
\vspace{-0.1cm}
\begin{equation}
    p_\theta(x_{t-k} \mid x_t, y) := q(x_{t-k} \mid x_t, \hat{x}_0 = G_{(x_t, y, t)})
\vspace{-0.1cm}
\label{eq:diff7}
\end{equation}
Meanwhile, for two adversarial diffusion processes, discriminators \(D_x\) and \(D_y\) respectively evaluate pairs (\(x_t\), \(\hat{x}_{t-k}\)) and (\(y_t\), \(\hat{y}_{t-k}\)), to determine their authenticity, as well as pairs (\(x_t\), \(x_{t-k}\)), (\(y_t\), \(y_{t-k}\)). After completing small reverse diffusion steps, the final denoised image can be sampled from the distribution \(\hat{x}_0 \sim p_\theta(x_0 \mid x_k, y)\). 

\begin{figure*}[tb]
    \centering
    \includegraphics[width=0.83\textwidth]{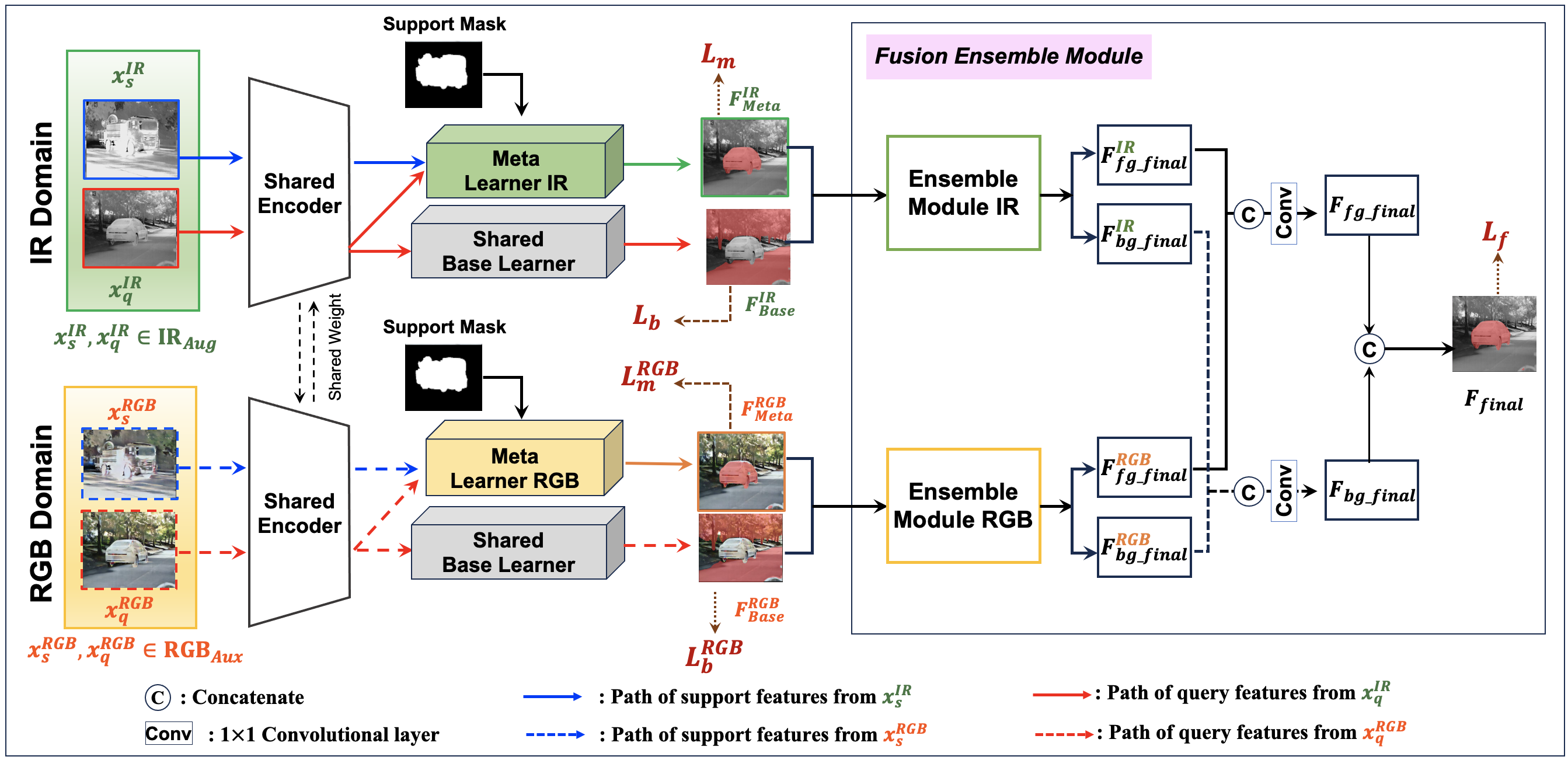}

    \vspace{-0.2cm}
    \caption{The proposed methods include two meta learners (Meta Learner IR and Meta Learner RGB), a shared base learner, and a fusion ensemble module. Each meta-learner evaluates the relationship between support and query in the IR and RGB domains, yielding \( F_{Meta}^{IR} \) and \( F_{Meta}^{RGB} \) respectively. The shared base learner produces predictions \( F_{Base}^{IR} \) and \( F_{Base}^{RGB} \). The fusion ensemble module integrates these predictions to generate final foreground and background probability maps \( F_{fg\_final} \) and \( F_{bg\_final} \), leading to the generation of \( F_{final} \).}
    \label{fig:msanet}

  \vspace{-0.3cm}
  
\end{figure*}

%%%%%%%%%% Methodology 
\section{Methodology}
To enhance the performance of FSS for IR data, we incorporate additional data derived from outputs of generative diffusion networks using fusion models. Our main motivation is to prioritize semantic preservation and enhance contrast or color information in IR images to improve the performance of IR FSS models. To this end, we leverage generative diffusion models to synthesize `Generated RGB Images' for auxiliary information, which are used in conjunction with novel fusion networks for integrating features from IR and these auxiliary data. Additionally, we also use `Generated Lightness Images' for data augmentation to address IR data scarcity issues.

\subsection{Few-Shot Segmentation Baseline} 
For completeness, we summarize the key features of the baseline network, MSANet~\cite{iqbal2022msanet}, which we use for FSS. MSANet, inspired by the BAM architecture~\cite{lang2022learning}, consist of three main parts including a base learner, a meta learner, and an ensemble module as illustrated in~\cref{fig:overview}(a). Further architectural and training details are provided in SuppMat.

The base learner is PSPNet~\cite{zhao2017pyramid}, with a ResNet-50/101~\cite{he2016deep}. Trained on known base classes in a supervised manner, it predicts regions of base classes in query images and suppresses falsely activated regions in the meta learner's output, yielding \( \mathit{F}_{\text{Base}} \). The base learner's loss is:
\vspace{-0.1cm}
\begin{equation}
    L_b = \frac{1}{N_{\text{batch}}} \sum_{i=1}^{N_{\text{batch}}} \text{CE}(\text{softmax}(\mathit{F}_{\text{Base}}^{(i)}), m_{\text{Base}}^{q(i)}),
\vspace{-0.1cm}
\label{eq:loss_base}
\end{equation}

\noindent where CE denotes the cross-entropy loss, \( m_{Base}^{q} \) represents the ground truth of the base classes in the query set, and \( N_{\text{batch}} \) is the batch size.

\looseness=-1
The meta learner aims to recognize unseen classes by discerning relationships between these features. Using intermediate to high-level features from the support set, the meta learner employs masked average pooling to create class prototype vectors and an attention module for foreground-focused feature mapping. It also generates a prior mask similar to PFENet~\cite{tian2020prior} and utilizes a multi-similarity module across multiple layers from both support and query sets to establish visual correspondences. The loss for the meta learner is calculated using binary cross-entropy (BCE):
\vspace{-0.1cm}
\begin{equation}
    L_m = \frac{1}{N_{\text{episode}}} \sum_{i=1}^{N_{\text{episode}}} \text{BCE}(\text{softmax}(\mathit{F}_{\text{Meta}}^{(i)}), m_{\text{Meta}}^{q(i)}),
\vspace{-0.1cm}
\label{eq:loss_meta}
\end{equation}

\noindent where \(\mathit{F}_{\text{Meta}}\) is the binary meta prediction mask, \(m_{\text{Meta}}^{q}\) are the labels for unseen classes in the query set, and \(N_{\text{episode}}\) is the total training episodes per batch.

   \begin{figure*}[t]
    \centering
    \includegraphics[width=0.85\textwidth]{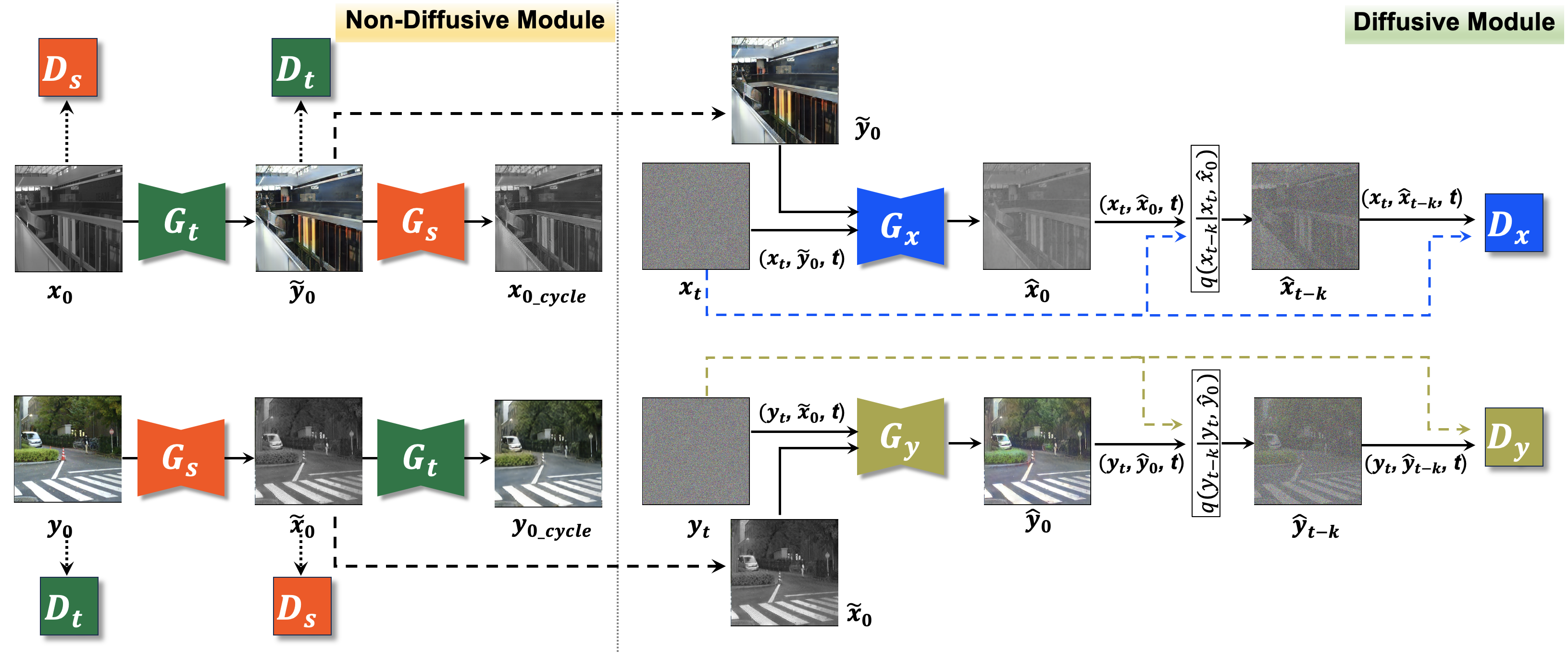}
    \vspace{-0.3cm}
    
    \caption{The adversarial conditional diffusion model~\cite{ozbey2023unsupervised} we adapt facilitates I2I translation between two different domains. The figure showcases examples of IR-RGB domain translation (Domain \(x\): IR, Domain \(y\): RGB). Non-diffusive modules (green and orange) employ two generator-discriminator pairs to provide initial estimates of the generated images for unpaired training with cycle-consistent loss~\cite{zhu2017unpaired}. Diffusive models (blue and yellow) utilize these initial estimates (black dashed lines) as conditioning for the denoising process.}

  \label{fig:syndiff}
  \vspace{-0.3cm}
\end{figure*}

The ensemble module enhances segmentation accuracy by refining predictions \( \mathit{F}_{\text{Meta}} \) and \( \mathit{F}_{\text{Base}} \) from both learners through an adjustment and an ensemble phase. In the former phase, it derives an adjustment factor map from the differences between query and support image pairs, and refines \( \mathit{F}_{\text{Meta}} \)'s foreground and background. In the latter phase, this adjusted background is combined with with \( \mathit{F}_{\text{Base}} \) to produce the final background map. These processes yield the final prediction map \( \mathit{F}_{\text{final}} \). The loss for final prediction is:
\vspace{-0.1cm}
\begin{equation}
    L_f = \frac{1}{N_{\text{episode}}} \sum_{i=1}^{N_{\text{episode}}} \text{BCE}(\text{softmax}(\mathit{F}_{\text{final}}^{(i)}), m_{\text{Meta}}^{q(i)}),
\vspace{-0.1cm}
\label{eq:loss_final}
\end{equation}

We designate this network as our baseline for the novel modifications outlined in~\cref{fig:msanet} and Section \ref{sec33}.

\subsection{Image-to-Image Translation} \label{sec32}
\begin{figure}[b]
    \centering
    \vspace{-.5cm}
    \includegraphics[width=0.8\columnwidth]{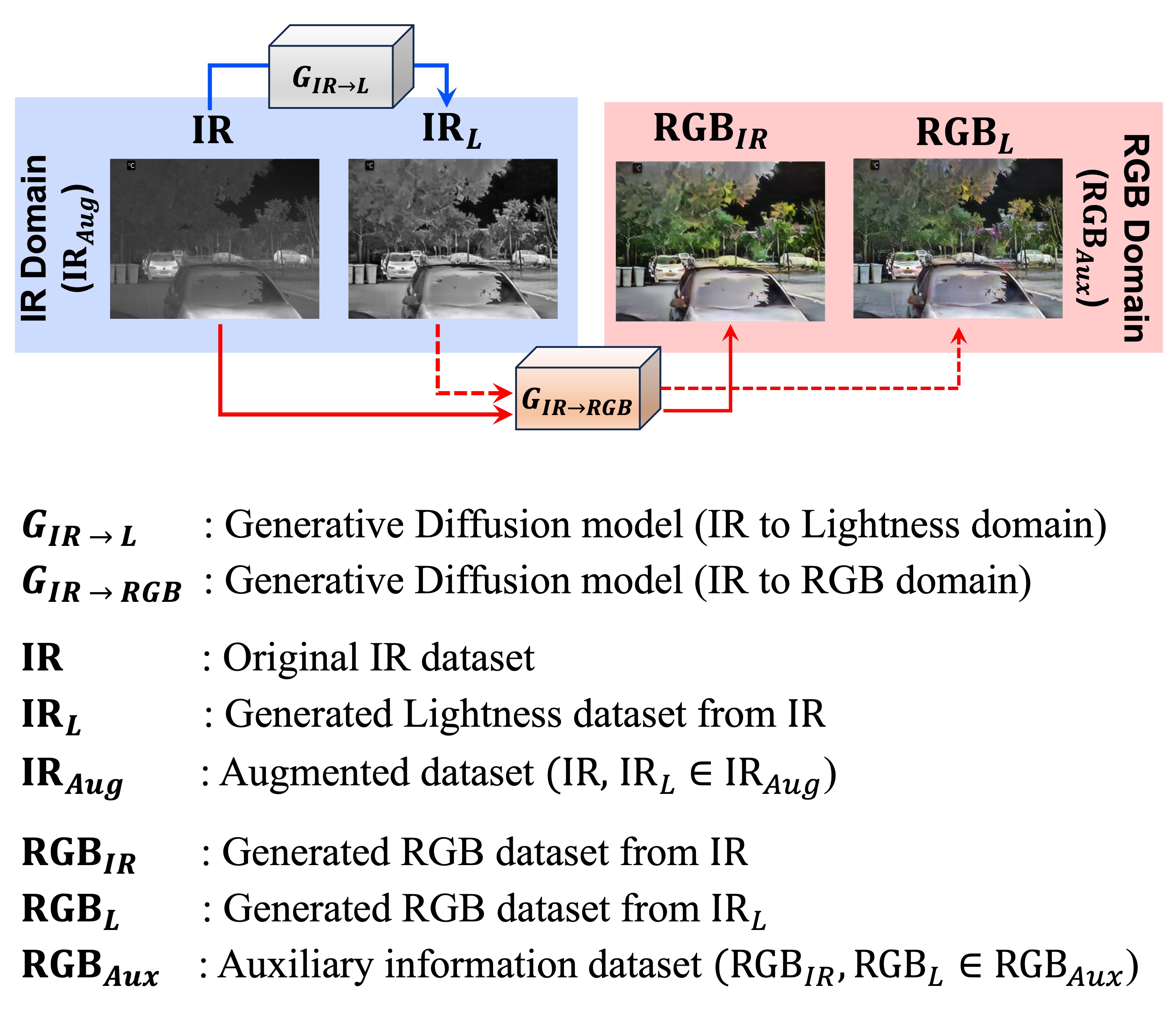}
    \vspace{-0.2cm}
    \caption{Description of the proposed generated datasets, including the naming conventions and purposes (augmentation versus auxiliary information).}
    \label{fig:datasets}
\end{figure}

We adopt the adversarial conditional diffusion model in \cite{ozbey2023unsupervised} as a generative network to perform I2I translation as shown in~\cref{fig:syndiff}. This model is trained using unpaired IR-RGB images from a publicly available IR-RGB dataset. Afterward, the trained model generates two types of new datasets: generated lightness dataset for data augmentation and generated RGB dataset for auxiliary information, derived from existing IR images, as outlined in~\cref{fig:datasets}.

\looseness=-1
\noindent \textbf{Data Augmentation (}$\textrm{IR}_{Aug}$\textbf{).} To mitigate the scarcity of IR data in public databases, we adapt a data augmentation strategy using generative models. RGB images from IR-RGB data are first converted to the LAB color space. We then employ unpaired I2I translation, where IR images are randomly selected without paired lightness domain (\textit{L}) images. This approach generates lightness images \(\textrm{IR}_{L}\) from existing IR images, preserving their grayscale properties while enhancing boundaries. This method effectively expands the training dataset for IR image-based tasks without requiring additional annotations, thereby diversifying data distributions and improving model robustness. We define the dataset \(\textrm{IR}_{Aug}\), which comprises both \(\textrm{IR}\) and \(\textrm{IR}_{L}\).

\vspace{2pt}

\noindent {\bf Auxiliary Information (}\(\textrm{RGB}_{Aux}\)\textbf{).} A significant innovation in our work is the use of generative models to complement the limited contrast in IR images, which complicates the adoption of existing FSS models designed for RGB images. This challenge primarily arises from spectral differences between IR and visible spectra. We hypothesize that enhancing the model with plausible colors to distinguish object boundaries and features could enhance the FSS process. To achieve this, we employ a generative model for unpaired I2I translation on the IR-RGB dataset, resulting in \(\textrm{RGB}_{IR} \).

Furthermore, we also propose to augment this strategy by converting previously obtained lightness images \(\textrm{IR}_{L} \) into the RGB domain, yielding \( \textrm{RGB}_L \). Thus, \( \textrm{RGB}_{IR} \) and \( \textrm{RGB}_L \) datasets serve as auxiliary information in our methodology, and is denoted as the dataset \( \textrm{RGB}_{Aux} \).
The major benefit of our approach is that we can pair augmented IR images \( \textrm{IR}_{aug} \), comprising both \( \textrm{IR} \) and \( \textrm{IR}_{L} \), with generated RGB datasets, \( \textrm{RGB}_{Aux} \). 
%------------------------------------------------------------------------
%------------------------------------------------------------------------

\subsection{Novel Fusion for Auxiliary Information} \label{sec33}
\noindent \textbf{Extracting Informative Features from (}\(\textrm{RGB}_{Aux}\)\textbf{).}
We propose a strategy employing an encoder and a novel component, the \textit{Meta Learner RGB}, which is structurally identical to the \textit{Meta Learner IR}, to leverage auxiliary information from the \(\textrm{RGB}_{Aux}\) dataset, as shown in~\cref{fig:msanet} and further detailed in SuppMat. Initially, the encoder is trained using both \(\textrm{IR}_{Aug}\) and \(\textrm{RGB}_{Aux}\) data to capture shared features across the IR and RGB domains during the base learner training stage. The base learner is shared across two domains and produces two probability maps: one from the IR domain query set and another from the RGB domain query set. Subsequently, utilizing features from the pre-trained encoder, Meta Learner RGB evaluates the similarity between support and query sets in RGB domain, while Meta Learner IR performs a similar task in IR domain, leveraging shared weights from atrous spatial pyramid pooling (ASPP) module~\cite{chen2014semantic} and the decoder. Consequently, this approach yields two meta predictions and two base predictions.

\vspace{2pt}

\noindent \textbf{Fusion for IR and RGB Domains.}
To harness these essential predictions, we propose a fusion ensemble module, as depicted in~\cref{fig:msanet}. This module comprises two identical ensemble modules and convolutional layers. Each ensemble module processes meta and base predictions separately in the IR and RGB domains, resulting in foreground/background predictions for each domain. Finally, foreground predictions from each domain are merged using \(1\times1\) convolutional layers, and similarly, background predictions are merged, resulting in final fore/background predictions. This fusion ensemble method effectively integrates features from different modalities. For the loss, we include \(L_{m}^{RGB}\) and \(L_{b}^{RGB}\), defined in an analogous manner to~\cref{eq:loss_base} and~\cref{eq:loss_meta}, yielding our proposed total loss:
\vspace{-0.1cm}
\begin{equation}
    L_{\text{total, proposed}} = L_{f} + L_{m} + L_{b} + L_{m}^{RGB} + L_{b}^{RGB}
\label{eq:loss_proposed}
\vspace{-0.1cm}
\end{equation}
Note our novel architectural changes only lead to a modest increase of 2.18\% and 1.60\% in total parameters over the baseline, using ResNet-50 and ResNet-101, respectively.

%%%%%%%%%% Experiments 
\section{Experiments}

\subsection{Few-Shot Segmentation Baseline} \label{sec41}
Since the IR FSS problem has not been extensively explored in the literature, there are no established baselines. We choose MSANet as our baseline FSS model and enhance its performance using generative model-based fusion techniques. We conduct experiments on two IR datasets, evaluating four methods: \texttt{baseline} with IR; \texttt{method1}, combining \(\textrm{IR}\) and simple augmentation-based learning with \(\textrm{IR}_{L}\); \texttt{method2} and \texttt{method3}, which integrate the baseline method with additional meta learners and fusion networks. \texttt{method2} uses \(\textrm{IR}\) and \(\textrm{RGB}_{IR}\), while \texttt{method3} utilizes the full set of proposed auxiliary data generated by the diffusion model \(\textrm{IR}_{Aug}\) and \(\textrm{RGB}_{Aux}\) data.

\subsection{Data}
\noindent {\bf IR Datasets.} To demonstrate the effectiveness of our approach, we utilize two IR image datasets: SODA~\cite{li2020segmenting} and SCUTSEG~\cite{xiong2021mcnet}. SODA dataset contains 2,168 IR images (1,168 for training, 1,000 for testing). It features labeled images of 20 semantic regions from diverse real-world scenes, encompassing both indoor and outdoor environments with varying lighting conditions, imaging blur, and resolutions. SCUTSEG dataset consists of 2,010 nighttime driving IR images (1,345 for training, 665 for testing) depicting common objects encountered during driving across 9 classes.

\looseness=-1
During the meta learner training stage, each dataset follows the FSS methodology~\cite{shaban2017one}, with object categories evenly divided into four folds, as shown in~\cref{tab:dataset}. FSS models are trained on three folds (unseen categories) and tested on the remaining fold (seen categories)  in a cross-validation manner. For validation, 1,000 pairs of support and query images are randomly selected from each fold. In SCUTSEG dataset, the `Road' class is designated as the background due to its presence in all images, making it unsuitable for the FSS problem setting. Both IR datasets are utilized into our FSS models and also converted to lightness and RGB images using generative models to enhance the FSS models with the proposed methods. Meanwhile, during the base learner training stage, the base learner is trained with three folds and sets the remaining fold  as the background, using the supervised protocol. Note that the two datasets could not be combined or used for domain switch testing due to differences in the labeling of overlapping classes, e.g. leaves are included in 'Tree' labels in SODA but not in SCUTSEG.

\begin{table}[t]
  \scriptsize
  \centering
  \setlength{\tabcolsep}{3pt}
  \renewcommand{\arraystretch}{1.0}
  \begin{tabular}{c|c|c|c}  
    \toprule
        \textbf{Fold} &   \textbf{SODA~\cite{li2020segmenting}} &   \textbf{Fold} &    \textbf{SCUTSEG~\cite{xiong2021mcnet}} \\  
       
    \midrule
      0 & Person, Building, Tree, Road, Pole          & 0 &  Person, Truck  \\
      1 & Grass, Door, Table, Chair, Car              & 1 &  Car, Pole      \\
      2 & Bicycle, Lamp, Monitor, Lane, Trash Can     & 2 &  Rider, Bus     \\
      3 & Animal, Fence, Sky, River, Side Walk        & 3 &  Fence, Tree    \\
    \bottomrule
    
  \end{tabular}
  \vspace{-0.2cm}
  \caption{Details of Dataset}
  \label{tab:dataset}
  \vspace{-0.5cm}
\end{table}

\vspace{2pt}
\noindent {\bf IR-RGB Datasets for Training the Adversarial Generative Diffusion Models.} 
We utilize two paired IR-RGB datasets, MFNet~\cite{ha2017mfnet} and RGB-NIR Scene~\cite{brown2011multi}, to pre-train the generative model. The MFNet dataset comprises 1,569 city view visual-thermal paired images, with 820 images captured during the daytime and 749 images captured at nighttime. To optimize contrast and improve I2I task performance, we exclude the nighttime RGB images due to their darkness, indistinct boundaries, and glare from artificial light, which pose challenges for I2I performance. However, we retain both daytime and nighttime IR images to enhance dataset versatility. The RGB-NIR Scene dataset includes 477 images across 9 categories, captured in both RGB and Near-IR. During training, we integrate both the MFNet and RGB-NIR datasets to train the generative model. For data augmentation, we employ unpaired IR-lightness I2I translation. Additionally, we utilize IR-RGB data for auxiliary purposes. Finally, the trained weights are applied to augment our IR datasets and generate auxiliary data.
\begin{table*}[t]
  \fontsize{13}{15}\selectfont  
  \centering
  \resizebox{\textwidth}{!}{%
  \renewcommand{\arraystretch}{1.00}
  \begin{tabular}{ c | c | c | c c c c | c c c c c c | c c c c c c}
  
    \toprule
        \multirow{2}{*}{\textbf{Backbone}} & \multirow{2}{*}{\textbf{Data}} & \multirow{2}{*}{\textbf{Method}} & \multirow{2}{*}{IR} & \multirow{2}{*}{IR$_{L}$} & \multirow{2}{*}{RGB$_{IR}$} & \multirow{2}{*}{RGB$_{L}$} & \multicolumn{6}{c}{\textbf{1-shot}} \vline & \multicolumn{6}{c}{\textbf{5-shot}} \\
        
        & & & & & & & Fold-0 & Fold-1 & Fold-2 & Fold-3 & \textbf{MIoU$\%$} & \textbf{FB-IoU$\%$} & Fold-0 & Fold-1 & Fold-2 & Fold-3 & \textbf{MIoU$\%$} & \textbf{FB-IoU$\%$} \\ 
   
    % ResNet-50  
    % SODA Results 
    \midrule
     \multirow{8}{*}{\textbf{ResNet-50}} & \multirow{4}{*}{\textbf{SODA}} 
     & \texttt{Baseline} & \checkmark & & & &
     43.58 & 37.35 & 47.45 & 55.34 & 45.93 & 70.15 & 
     48.15 & 42.00 & 61.23 & 59.39 & 52.69 & 74.44 \\ 

     % Method1
     & & \texttt{Method1} & \checkmark & \checkmark & & &
     44.26 & 38.03 & 48.29 & 60.10 & \underline{47.67} & \underline{71.14} & 
     51.15 & 41.29 & 59.75 & 63.89 & 54.02 & \underline{75.48}  \\

     % Method2
     & & \texttt{Method2} & \checkmark & & \checkmark & &
     41.92 & 37.75 & 51.10 & 58.51 & 47.32 & 70.35 & 
     49.21 & 41.05 & 62.95 & 64.42 & \underline{54.41} & 75.23 \\
     
     % Method3
     & & \texttt{Method3} & \checkmark & \checkmark & \checkmark & \checkmark &
     44.01 & 38.48 & 50.60 & 61.19 & \textbf{48.57} & \textbf{71.81} & 
     50.92 & 42.79 & 60.09 & 65.36 & \textbf{54.79} & \textbf{75.98} \\ \cline{2-19}

     % SCUTSEG Results
     & \multirow{4}{*}{\textbf{SCUTSEG}} 
     & \texttt{Baseline} & \checkmark &  &  &  & 
     48.35 & 27.29 & 46.51 & 3.97 & 31.53 & 66.68 & 
     51.69 & 35.46 & 48.53 & 13.20 & 37.22 & 68.68 \\ 

     % Method1
     & & \texttt{Method1} & \checkmark & \checkmark &  &  & 
     54.25 & 34.22 & 48.84 & 14.26 & \underline{37.89} & \underline{69.51} & 
     56.69 & 43.88 & 53.03 & 13.25 & 41.71 & \underline{71.67}  \\

     % Method2
     & & \texttt{Method2} & \checkmark &  & \checkmark & & 
     50.39 & 30.85 & 51.29 & 12.11 & 36.16 & 68.25 & 
     55.38 & 39.98 & 52.29 & 21.14 & \underline{42.20} & 70.56 \\

     % Method3
     &  & \texttt{Method3} & \checkmark & \checkmark & \checkmark &  \checkmark &  
     55.44 & 34.30 & 55.09 & 16.51 & \textbf{40.33} & \textbf{70.48} & 
     57.46 & 41.75 & 54.29 & 28.00 & \textbf{45.38} & \textbf{72.43} \\ \cline{1-19}

    % ResNet-101      
     \multirow{8}{*}{\textbf{ResNet-101}} & \multirow{4}{*}{\textbf{SODA}} 

     & \texttt{Baseline} & \checkmark & & & &
     42.39 & 38.73 & 53.59 & 58.11 & 48.20 & 70.72 & 
     48.20 & 42.72 & 64.34 & 63.88 & 54.78 & 75.18 \\ 

     % Method1
     & & \texttt{Method1} & \checkmark & \checkmark & & & 
     43.74 & 39.04 & 52.89 & 61.93 & 48.54 & \underline{72.13} & 
     51.41 & 43.45 & 60.18 & 66.27 & 55.33 & \textbf{76.30} \\

     % Method2
     & & \texttt{Method2} & \checkmark & & \checkmark & &
     42.40 & 38.56 & 52.89 & 61.93 & \underline{48.94} & 71.58 & 
     50.19 & 43.63 & 63.81 & 65.00 & \underline{55.66} & 75.17 \\

     % Method3
     & & \texttt{Method3} & \checkmark & \checkmark & \checkmark &  \checkmark &  
     45.38 & 38.04 & 52.51 & 60.90 & \textbf{49.21} & \textbf{72.21} & 
     51.96 & 42.50 & 65.64 & 65.74 & \textbf{56.46} & \underline{76.14} \\ \cline{2-19}
     
    % SCUTSEG Results 
     % \hline
     & \multirow{4}{*}{\textbf{SCUTSEG}} 
     & \texttt{Baseline} & \checkmark &  &  &  &
     50.38 & 30.18 & 44.68 & 11.83 & 34.27 & 66.96 & 
     52.48 & 39.26 & 47.18 & 21.75 & 40.17 & 68.49 \\ 

     % Method1
     & & \texttt{Method1} & \checkmark & \checkmark &  &  &
     56.08 & 34.90 & 50.70 & 13.28 & 38.74 & 69.29 & 
     61.61 & 43.91 & 56.36 & 22.18 & 46.01 & 72.73  \\

     % Method2
     & & \texttt{Method2} & \checkmark &  & \checkmark &  &
     59.15 & 36.50 & 51.69 & 14.88 & \underline{40.56} & \underline{70.68} & 
     61.97 & 45.95 & 56.76 & 25.50 & \underline{47.55} & \underline{73.14} \\

     % Method3     
     & & \texttt{Method3} & \checkmark & \checkmark & \checkmark &  \checkmark &  
     62.36 & 37.11 & 52.91 & 14.45 & \textbf{41.71} & \textbf{71.14} & 
     66.49 & 46.83 & 56.03 & 25.90 & \textbf{48.81} & \textbf{73.62} \\
     \bottomrule
     
  \end{tabular}
  }
  \vspace{-.2cm}
  \caption{Comparison of each method on SODA~\cite{li2020segmenting} and SCUTSEG~\cite{xiong2021mcnet} datasets under 1-shot and 5-shot settings. (Upper table: ResNet-50, lower table: ResNet-101.) Entries in \textbf{bold} indicate the best performance, while those \underline{underlined} denote the second best.}
\vspace{-.3cm}
  \label{tab:proposed_results}
\end{table*}

\subsection{Implementation Details}
We follow the basic training settings for both MSANet~\cite{iqbal2022msanet} and SynDiff~\cite{ozbey2023unsupervised}, with a few modifications.
\vspace{3pt}

\noindent {\bf Few-Shot Segmentation Model.} The training of MSANet is divided into two stages. First, the base learner is optimized on each fold of the IR dataset for 200 epochs using PSPNet~\cite{zhao2017pyramid} with a pre-trained ResNet-50/101~\cite{he2016deep} backbone. The encoder is trained with all data designated for each method to capture features across both IR and RGB domains as mentioned in Section~\ref{sec41}. Second, the meta learner and ensemble module are trained in the episodic training paradigm~\cite{vinyals2016matching} for 200 epochs, stopping if no improvement in validation set results occurs for 75 epochs. Both learners use the SGD optimizer with a learning rate of $7.5 \cdot 10^{-3}$. Input images are cropped to 473$\times$473, with standard augmentations during training. 

Evaluation is based on averaging results from 5 testing runs with different random seeds, using mean Intersection-over-Union (mIoU) and foreground-background IoU (FB-IoU) metrics, as in previous FSS works.

\vspace{2pt}
\noindent {\bf Adversarial Generative Diffusion Model.} We train the generative model from scratch, setting as \(T=1000\), a step size of \(k=250\), and \(T/k=4\) diffusion steps. The training of the adversarial model in~\cite{ozbey2023unsupervised} is designed for medical data with single input and output channels. Therefore, when generating lightness images, we follow this configuration, while for generating RGB images, we modify the input and output channels to 3. Additionally, we set the weight for cycle-consistency from 0.5 to 10 to preserve characteristics captured in different spectra without altering contrast. We set 100 epochs for I2I from IR to RGB domains and 50 epochs for I2I from IR to lightness domains. Input images, which are unpaired, are scaled to \(256\times256\). We apply gamma correction and histogram equalization~\cite{zuiderveld1994contrast} on IR dataset to effectively translate the IR domain into other domains, introducing non-linearity and enhancing contrast.

 \begin{figure*}[t]
    \centering
    \includegraphics[width=15.0cm, height=8.5cm]{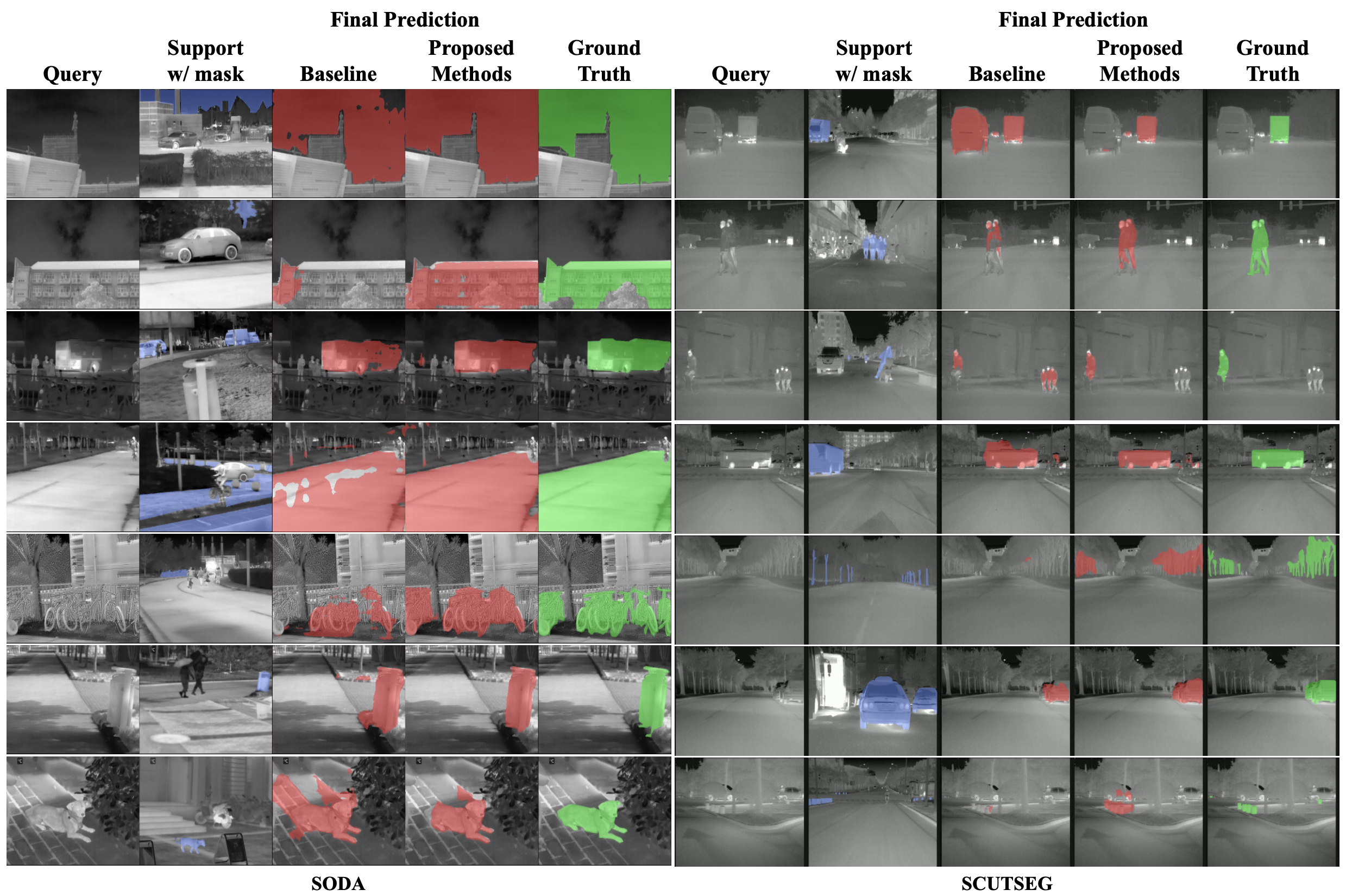}
    \vspace{-.2cm}
    \caption{The qualitative evaluation of baseline and the proposed method (\texttt{method3}) on SODA and SCUTSEG under 1-shot setting.}
    \label{fig:qual_msanet}
    \vspace{-.3cm}
\end{figure*}

\subsection{Evaluation of Proposed FSS Methods}
~\cref{tab:proposed_results} illustrates the performance of different approaches on SODA and SCUTSEG datasets. We experiment with ResNet-50 and ResNet-101 as encoders for versatility, and further ablation studies are provided in SuppMat. In \cref{tab:SOTA}, we report a comparison of our proposed method with recent RGB-based SOTA FSS approaches, including PFENet~\cite{tian2020prior}, HSNet~\cite{min2021hypercorrelation}, BAM~\cite{lang2022learning}, VAT~\cite{hong2022cost}, MSANet~\cite{iqbal2022msanet} and MSI~\cite{moon2023msi} on IR datasets. Qualitative evaluations of the generative diffusion models, as well as implementation details for SOTA methods are available in SuppMat.
Note detailed quantitative analysis of the generative model was not prioritized in our study, as the main advantage of our novel approach lies in the multi-model fusion model’s ability to distill features from synthetic RGB data for improved segmentation without focusing on the quality of the I2I translation task.
\begin{table}[t]
  \fontsize{6.3}{8}\selectfont  
  \centering
  \setlength{\tabcolsep}{1.7pt}
  \renewcommand{\arraystretch}{0.93}
  \begin{tabular}{ c | >{\raggedleft\arraybackslash}p{1.5cm} | c c | c c | c c | c c}
  
    \toprule
        \multirow{3}{*}{\textbf{Data}} & \multirow{3}{*}{\textbf{Model}} & \multicolumn{4}{c}{\textbf{ResNet-50}} \vline & \multicolumn{4}{c}{\textbf{ResNet-101}} \\ \cline{3-10}
        & & \multicolumn{2}{c}{\textbf{1-shot}} \vline & \multicolumn{2}{c}{\textbf{5-shot}} \vline  & \multicolumn{2}{c}{\textbf{1-shot}} \vline & \multicolumn{2}{c}{\textbf{5-shot}} \\ 
        & & \textbf{MIoU} & \textbf{FB-IoU} 
        & \textbf{MIoU} & \textbf{FB-IoU} 
        & \textbf{MIoU} & \textbf{FB-IoU} 
        & \textbf{MIoU} & \textbf{FB-IoU}\\
    \midrule
     \multirow{7}{*}{{\rotatebox{90}{\textbf{SODA~\cite{li2020segmenting}}}}} 

     % PFENet
     & PFENet~\cite{tian2020prior} &
     30.36 & 55.00 & 
     37.49 & 61.30 & 
     36.51 & 58.72 & 
     45.91 & 66.63  \\ 

     % HSNet
     &  HSNet~\cite{min2021hypercorrelation} &
     31.67 & 58.43 & 
     40.73 & 64.25 & 
     35.69 & 60.69 & 
     44.99 & 66.50  \\ 

     % BAM
     & BAM~\cite{lang2022learning} &
     41.56 & 65.94 & 
     49.40 & 71.77 & 
     45.37 & 68.26 & 
     52.05 & 73.68  \\ 

     % VAT
     & VAT~\cite{hong2022cost} &
     34.01 & 59.58 & 
     40.95 & 63.39 & 
     37.85 & 62.33 & 
     45.90 & 67.20  \\ 

     % MSANet
     & MSANet~\cite{iqbal2022msanet} &
     \underline{45.93} & \underline{70.15} & 
     \underline{52.69} & \underline{74.44} &     
     \underline{48.20} & \underline{70.72} & 
     \underline{54.78} & \underline{75.18} \\   

     % MSI
     & MSI~\cite{moon2023msi} &
     34.33 & 59.22 & 
     40.76 & 62.75 & 
     36.83 & 61.50 & 
     43.44 & 65.57    \\ \cline{2-10}

     \rowcolor{gray!20!white}
     \cellcolor{white} & Ours (Method3) &
     \textbf{48.57} & \textbf{71.81} & 
     \textbf{54.79} & \textbf{75.98} &     
     \textbf{49.21} & \textbf{72.21} & 
     \textbf{56.46} & \textbf{76.14}  \\ 
     
      %%%%%%% SCUTSEG Results %%%%%%%%
     \hline
     \multirow{7}{*}{{\rotatebox{90}{\textbf{SCUTSEG~\cite{xiong2021mcnet}}}}} 
     % PFENet
     & PFENet~\cite{tian2020prior} &
     28.52 & 63.52 & 
     30.63 & 65.68 & 
     31.46 & 65.06 & 
     37.62 & 67.37  \\ 

     % HSNet
     &  HSNet~\cite{min2021hypercorrelation} &
     24.95 & 61.30 & 
     29.93 & 64.62 & 
     24.11 & 61.38 & 
     30.29 & 64.71  \\ 

     % BAM
     & BAM~\cite{lang2022learning} &
     29.92 & 65.93 & 
     33.18 & 67.91 & 
     32.46 & 65.35 & 
     \underline{40.76} & \underline{69.04}  \\ 

     % VAT
     & VAT~\cite{hong2022cost} &
     26.13 & 61.36 & 
     30.42 & 63.67 & 
     27.61 & 62.55 & 
     33.08 & 65.12  \\ 

     % MSANet
     & MSANet~\cite{iqbal2022msanet} &

     \underline{31.53} & \underline{66.68} & 
     \underline{37.22} & \underline{68.68} &      
     \underline{34.27} & \underline{66.96} & 
     40.17 & 68.49 \\

     % MSI
     & MSI~\cite{moon2023msi} &
     26.06 & 62.54 & 
     28.32 & 63.93 & 
     25.59 & 62.03 &   
     27.55 & 63.61  \\ \cline{2-10}

     \rowcolor{gray!20!white}
     \cellcolor{white} & Ours (Method3) &
     \textbf{40.33} & \textbf{70.48} & 
     \textbf{45.38} & \textbf{72.43} &
     \textbf{41.71} & \textbf{71.14} & 
     \textbf{48.81} & \textbf{73.62} \\ 
     
    \bottomrule
    
  \end{tabular}
  
  \vspace{-.2cm}
  \caption{Comparison with SOTA methods on IR datasets. The results indicate the average mIoU and FB-IoU across four folds.}
  
\vspace{-.6cm}
  \label{tab:SOTA}
\end{table}

\noindent \textbf{Quantitative Results.} Our proposed methods consistently outperform the baseline, with notable improvements in both datasets as shown in~\cref{tab:proposed_results}. Results with ResNet-101 improve on ResNet-50 in both datasets. Additionally, the 5-shot setting outperforms the 1-shot setting, consistent with previous reports on five class representatives in the 5-shot setting allowing the FSS model to better understand the target class compared to the 1-shot setting~\cite{tian2020prior}.

Specifically, the inclusion of the augmentation method (\texttt{method1}) and the integration of an additional meta-learner and fusion ensemble module with auxiliary data (\texttt{method2}), significantly enhance both mIoU and FB-IoU compared to the baseline in all scenarios. Further, \texttt{method3} yields 
the highest mIoU, with enhancements between 1.01\% to 2.64\% in SODA and 2.80\% to 8.64\% in SCUTSEG. \texttt{method3} also outperforms the other methods in terms of FB-IoU, with improvements between 0.96\% to 1.66\% in SODA and between 3.75\% to 5.13\% in SCUTSEG, except for the results in the 5-shot setting with ResNet-101 in SCUTSEG dataset. 

\noindent \textbf{Comparison with SOTA Methods.}
As depicted in~\cref{tab:SOTA}, our proposed method achieves the best performance compared to SOTA methods in terms of mIoU across all conditions, with improvements ranging from 1.01\% to 18.21\% in SODA dataset and from 7.44\% to 21.26\% in SCUTSEG dataset. It also demonstrates the highest FB-IoU.

\begin{figure}[b]
    \centering
    \vspace{-.7cm}
    \includegraphics[width=0.98\columnwidth]{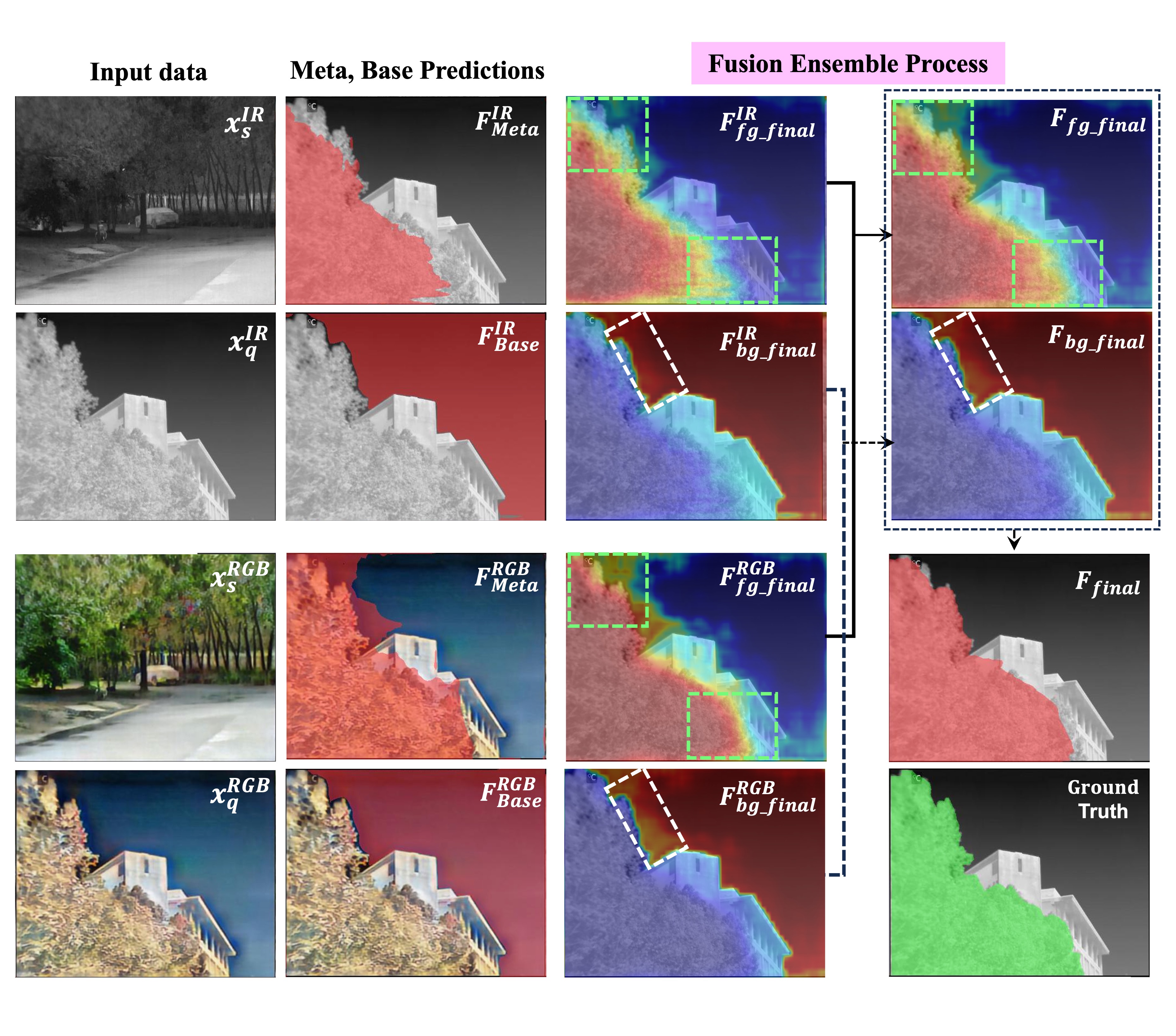}
    \vspace{-0.7cm}
    \caption{Fusion ensemble networks enhance FSS performance (notations as in~\cref{fig:msanet}).
    In the fusion ensemble process, higher values are mapped to red and lower values to blue.
    }
    \label{fig:fusion_ensemble}
\end{figure}

\noindent \textbf{Examples and Fusion Performance.} As illustrated in~\cref{fig:qual_msanet}, \texttt{method3} outperforms the \texttt{baseline} in both datasets. This suggests that encoders trained with \(\textrm{IR}_{Aug}\) and \(\textrm{RGB}_{Aux}\) capture superior features from the query and support sets, and the additional meta learner and fusion ensemble networks enable FSS models to predict more accurately. Furthermore,~\cref{fig:fusion_ensemble} shows that the meta learner for the RGB domain emphasizes relationships within color spaces, while the fusion ensemble bridges information gaps in IR images by combining features from both IR and RGB domains (green box). The shared base learner further captures better similarity in IR than in RGB, and the fusion ensemble module integrates these effectively (white box), yielding precise final predictions.

%%%%%%%%%% Conclusion 
\section{Conclusion}
In this paper, we explore FSS of IR images, which holds significant promise and importance for various fields. FSS tackles the challenge of limited IR data for supervised training and the emergence of new classification classes. Our proposed techniques and results demonstrate the potential of using generative DL models and fusion ensemble networks to enhance FSS performance for IR images. Specifically, we use generative models for data augmentation and more critically to generate auxiliary information to tackle issues related to limited contrast in IR images. For the latter, we introduce a novel fusion ensemble module to use information from auxiliary generated data. Our methods improve FSS performance without requiring paired IR-RGB datasets as previous methods did.

%%%%%%%%%% Acknowledgements 
\section{Acknowledgements}
This work was partially supported by NIH R01EB032830, NIH P41EB027061.

%%%%%%%%%% Reference 
{\small
\bibliographystyle{ieee_fullname}
\bibliography{egbib}

\begin{thebibliography}{10}\itemsep=-1pt

\bibitem{alcalar2024ECCV}
Ya{\c{s}}ar~Utku Al{\c{c}}alar and Mehmet Ak{\c{c}}akaya.
\newblock Zero-shot adaptation for approximate posterior sampling of diffusion models in inverse problems.
\newblock In {\em Computer Vision -- ECCV 2024}, pages 444--460. Springer Nature, 2024.

\bibitem{bao2021visible}
Yanqi Bao, Kechen Song, Jie Wang, Liming Huang, Hongwen Dong, and Yunhui Yan.
\newblock Visible and thermal images fusion architecture for few-shot semantic segmentation.
\newblock {\em Journal of Visual Communication and Image Representation}, 80:103306, 2021.

\bibitem{bhattarai2020deep}
Manish Bhattarai and Manel Martinez-Ramon.
\newblock A deep learning framework for detection of targets in thermal images to improve firefighting.
\newblock {\em IEEE Access}, 8:88308--88321, 2020.

\bibitem{brown2011multi}
Matthew Brown and Sabine S{\"u}sstrunk.
\newblock Multi-spectral sift for scene category recognition.
\newblock In {\em CVPR 2011}, pages 177--184. IEEE, 2011.

\bibitem{bustos2023systematic}
Nicolas Bustos, Mehrsa Mashhadi, Susana~K Lai-Yuen, Sudeep Sarkar, and Tapas~K Das.
\newblock A systematic literature review on object detection using near infrared and thermal images.
\newblock {\em Neurocomputing}, page 126804, 2023.

\bibitem{chen2014semantic}
Liang-Chieh Chen, George Papandreou, Iasonas Kokkinos, Kevin Murphy, and Alan~L Yuille.
\newblock Semantic image segmentation with deep convolutional nets and fully connected crfs.
\newblock {\em arXiv preprint arXiv:1412.7062}, 2014.

\bibitem{danaci2023survey}
Kevser~Irem Danaci and Erdem Akagunduz.
\newblock A survey on infrared image \& video sets.
\newblock {\em Multimedia Tools and Applications}, pages 1--39, 2023.

\bibitem{dhariwal2021diffusion}
Prafulla Dhariwal and Nichol Alexander.
\newblock Diffusion models beat gans on image synthesis.
\newblock {\em Advances in neural information processing systems}, 34:8780--8794, 2021.

\bibitem{dong2018few}
Nanqing Dong and Eric~P Xing.
\newblock Few-shot semantic segmentation with prototype learning.
\newblock In {\em BMVC}, volume~3, 2018.

\bibitem{d2019cnn}
Antoine d’Acremont, Ronan Fablet, Alexandre Baussard, and Guillaume Quin.
\newblock Cnn-based target recognition and identification for infrared imaging in defense systems.
\newblock {\em Sensors}, 19(9):2040, 2019.

\bibitem{goodfellow2014generative}
Ian Goodfellow, Jean Pouget-Abadie, Mehdi Mirza, Bing Xu, David Warde-Farley, Sherjil Ozair, Aaron Courville, and Yoshua Bengio.
\newblock Generative adversarial nets.
\newblock {\em Advances in neural information processing systems}, 27, 2014.

\bibitem{goodfellow2020generative}
Ian Goodfellow, Jean Pouget-Abadie, Mehdi Mirza, Bing Xu, David Warde-Farley, Sherjil Ozair, Aaron Courville, and Yoshua Bengio.
\newblock Generative adversarial networks.
\newblock {\em Communications of the ACM}, 63(11):139--144, 2020.

\bibitem{ha2017mfnet}
Qishen Ha, Kohei Watanabe, Takumi Karasawa, Yoshitaka Ushiku, and Tatsuya Harada.
\newblock Mfnet: Towards real-time semantic segmentation for autonomous vehicles with multi-spectral scenes.
\newblock In {\em 2017 IEEE/RSJ International Conference on Intelligent Robots and Systems (IROS)}, pages 5108--5115. IEEE, 2017.

\bibitem{he2016deep}
Kaiming He, Xiangyu Zhang, Shaoqing Ren, and Jian Sun.
\newblock Deep residual learning for image recognition.
\newblock In {\em Proceedings of the IEEE Conference on Computer Vision and Pattern Recognition}, pages 770--778, 2016.

\bibitem{ho2020denoising}
Jonathan Ho, Ajay Jain, and Pieter Abbeel.
\newblock Denoising diffusion probabilistic models.
\newblock {\em Advances in neural information processing systems}, 33:6840--6851, 2020.

\bibitem{hong2022cost}
Sunghwan Hong, Seokju Cho, Jisu Nam, Stephen Lin, and Seungryong Kim.
\newblock Cost aggregation with 4d convolutional swin transformer for few-shot segmentation.
\newblock In {\em European Conference on Computer Vision}, pages 108--126. Springer, 2022.

\bibitem{hu2019attention}
Tao Hu, Pengwan Yang, Chiliang Zhang, Gang Yu, Yadong Mu, and Cees~GM Snoek.
\newblock Attention-based multi-context guiding for few-shot semantic segmentation.
\newblock In {\em Proceedings of the AAAI conference on artificial intelligence}, volume~33, pages 8441--8448, 2019.

\bibitem{iqbal2022msanet}
Ehtesham Iqbal, Sirojbek Safarov, and Seongdeok Bang.
\newblock Msanet: Multi-similarity and attention guidance for boosting few-shot segmentation.
\newblock {\em arXiv preprint arXiv:2206.09667}, 2022.

\bibitem{khellal2018convolutional}
Atmane Khellal, Hongbin Ma, and Qing Fei.
\newblock Convolutional neural network based on extreme learning machine for maritime ships recognition in infrared images.
\newblock {\em Sensors}, 18(5):1490, 2018.

\bibitem{kim2015real}
Jong-Hwan Kim and Brian~Y Lattimer.
\newblock Real-time probabilistic classification of fire and smoke using thermal imagery for intelligent firefighting robot.
\newblock {\em Fire Safety Journal}, 72:40--49, 2015.

\bibitem{koch2015siamese}
Gregory Koch, Richard Zemel, and Ruslan Salakhutdinov.
\newblock Siamese neural networks for one-shot image recognition.
\newblock In {\em ICML deep learning workshop}, volume~2. Lille, 2015.

\bibitem{kutuk2022semantic}
Z{\"u}lfiye K{\"u}t{\"u}k and G{\"o}rkem Algan.
\newblock Semantic segmentation for thermal images: A comparative survey.
\newblock In {\em Proceedings of the IEEE/CVF Conference on Computer Vision and Pattern Recognition}, pages 286--295, 2022.

\bibitem{lang2022learning}
Chunbo Lang, Gong Cheng, Binfei Tu, and Junwei Han.
\newblock Learning what not to segment: A new perspective on few-shot segmentation.
\newblock In {\em Proceedings of the IEEE/CVF conference on computer vision and pattern recognition}, pages 8057--8067, 2022.

\bibitem{li2020segmenting}
Chenglong Li, Wei Xia, Yan Yan, Bin Luo, and Jin Tang.
\newblock Segmenting objects in day and night: Edge-conditioned cnn for thermal image semantic segmentation.
\newblock {\em IEEE Transactions on Neural Networks and Learning Systems}, 32(7):3069--3082, 2020.

\bibitem{liu2023multi}
Jinyuan Liu, Zhu Liu, Guanyao Wu, Long Ma, Risheng Liu, Wei Zhong, Zhongxuan Luo, and Xin Fan.
\newblock Multi-interactive feature learning and a full-time multi-modality benchmark for image fusion and segmentation.
\newblock In {\em Proceedings of the IEEE/CVF international conference on computer vision}, pages 8115--8124, 2023.

\bibitem{min2021hypercorrelation}
Juhong Min, Dahyun Kang, and Minsu Cho.
\newblock Hypercorrelation squeeze for few-shot segmentation.
\newblock In {\em Proceedings of the IEEE/CVF international conference on computer vision}, pages 6941--6952, 2021.

\bibitem{moon2023msi}
Seonghyeon Moon, Samuel~S Sohn, Honglu Zhou, Sejong Yoon, Vladimir Pavlovic, Muhammad~Haris Khan, and Mubbasir Kapadia.
\newblock Msi: Maximize support-set information for few-shot segmentation.
\newblock In {\em Proceedings of the IEEE/CVF International Conference on Computer Vision}, pages 19266--19276, 2023.

\bibitem{ozbey2023unsupervised}
Muzaffer {\"O}zbey, Onat Dalmaz, Salman~UH Dar, Hasan~A Bedel, {\c{S}}aban {\"O}zturk, Alper G{\"u}ng{\"o}r, and Tolga {\c{C}}ukur.
\newblock Unsupervised medical image translation with adversarial diffusion models.
\newblock {\em IEEE Transactions on Medical Imaging}, 2023.

\bibitem{panetta2021ftnet}
Karen Panetta, KM~Shreyas Kamath, Srijith Rajeev, and Sos~S Agaian.
\newblock Ftnet: Feature transverse network for thermal image semantic segmentation.
\newblock {\em IEEE Access}, 9:145212--145227, 2021.

\bibitem{rakelly2018few}
Kate Rakelly, Evan Shelhamer, Trevor Darrell, Alexei~A Efros, and Sergey Levine.
\newblock Few-shot segmentation propagation with guided networks.
\newblock {\em arXiv preprint arXiv:1806.07373}, 2018.

\bibitem{sasaki2021unit}
Hiroshi Sasaki, Chris~G Willcocks, and Toby~P Breckon.
\newblock Unit-ddpm: Unpaired image translation with denoising diffusion probabilistic models.
\newblock {\em arXiv preprint arXiv:2104.05358}, 2021.

\bibitem{shaban2017one}
Amirreza Shaban, Shray Bansal, Zhen Liu, Irfan Essa, and Byron Boots.
\newblock One-shot learning for semantic segmentation.
\newblock {\em arXiv preprint arXiv:1709.03410}, 2017.

\bibitem{snell2017prototypical}
Jake Snell, Kevin Swersky, and Richard Zemel.
\newblock Prototypical networks for few-shot learning.
\newblock {\em Advances in neural information processing systems}, 30, 2017.

\bibitem{song2020denoising}
Jiaming Song, Chenlin Meng, and Stefano Ermon.
\newblock Denoising diffusion implicit models.
\newblock {\em arXiv preprint arXiv:2010.02502}, 2020.

\bibitem{song2020score}
Yang Song, Jascha Sohl-Dickstein, Diederik~P Kingma, Abhishek Kumar, Stefano Ermon, and Ben Poole.
\newblock Score-based generative modeling through stochastic differential equations.
\newblock {\em arXiv preprint arXiv:2011.13456}, 2020.

\bibitem{sun2019rtfnet}
Yuxiang Sun, Weixun Zuo, and Ming Liu.
\newblock Rtfnet: Rgb-thermal fusion network for semantic segmentation of urban scenes.
\newblock {\em IEEE Robotics and Automation Letters}, 4(3):2576--2583, 2019.

\bibitem{sun2020fuseseg}
Yuxiang Sun, Weixun Zuo, Peng Yun, Hengli Wang, and Ming Liu.
\newblock Fuseseg: Semantic segmentation of urban scenes based on rgb and thermal data fusion.
\newblock {\em IEEE Transactions on Automation Science and Engineering}, 18(3):1000--1011, 2020.

\bibitem{sung2018learning}
Flood Sung, Yongxin Yang, Li Zhang, Tao Xiang, Philip~HS Torr, and Timothy~M Hospedales.
\newblock Learning to compare: Relation network for few-shot learning.
\newblock In {\em Proceedings of the IEEE conference on computer vision and pattern recognition}, pages 1199--1208, 2018.

\bibitem{tian2020prior}
Zhuotao Tian, Hengshuang Zhao, Michelle Shu, Zhicheng Yang, Ruiyu Li, and Jiaya Jia.
\newblock Prior guided feature enrichment network for few-shot segmentation.
\newblock {\em IEEE transactions on pattern analysis and machine intelligence}, 44(2):1050--1065, 2020.

\bibitem{vinyals2016matching}
Oriol Vinyals, Charles Blundell, Timothy Lillicrap, and Daan Wierstra.
\newblock Matching networks for one shot learning.
\newblock {\em Advances in neural information processing systems}, 29, 2016.

\bibitem{wang2015infrared}
Bin Wang, Lili Dong, Ming Zhao, Houde Wu, Yuanyuan Ji, and Wenhai Xu.
\newblock An infrared maritime target detection algorithm applicable to heavy sea fog.
\newblock {\em Infrared Physics \& Technology}, 71:56--62, 2015.

\bibitem{wang2019panet}
Kaixin Wang, Jun~Hao Liew, Yingtian Zou, Daquan Zhou, and Jiashi Feng.
\newblock Panet: Few-shot image semantic segmentation with prototype alignment.
\newblock In {\em proceedings of the IEEE/CVF international conference on computer vision}, pages 9197--9206, 2019.

\bibitem{wang2018understanding}
Panqu Wang, Pengfei Chen, Ye Yuan, Ding Liu, Zehua Huang, Xiaodi Hou, and Garrison Cottrell.
\newblock Understanding convolution for semantic segmentation.
\newblock In {\em 2018 IEEE winter conference on applications of computer vision (WACV)}, pages 1451--1460. Ieee, 2018.

\bibitem{xiao2022tackling}
Zhisheng Xiao, Karsten Kreis, and Arash Vahdat.
\newblock Tackling the generative learning trilemma with denoising diffusion gans.
\newblock In {\em International Conference on Learning Representations}, 2022.

\bibitem{xiong2021mcnet}
Haitao Xiong, Wenjie Cai, and Qiong Liu.
\newblock Mcnet: Multi-level correction network for thermal image semantic segmentation of nighttime driving scene.
\newblock {\em Infrared Physics \& Technology}, 113:103628, 2021.

\bibitem{yang2022image}
Suorong Yang, Weikang Xiao, Mengcheng Zhang, Suhan Guo, Jian Zhao, and Furao Shen.
\newblock Image data augmentation for deep learning: A survey.
\newblock {\em arXiv preprint arXiv:2204.08610}, 2022.

\bibitem{yun2019improved}
Kyongsik Yun, Kevin Yu, Joseph Osborne, Sarah Eldin, Luan Nguyen, Alexander Huyen, and Thomas Lu.
\newblock Improved visible to ir image transformation using synthetic data augmentation with cycle-consistent adversarial networks.
\newblock In {\em Pattern Recognition and Tracking XXX}, volume 10995, page 1099502. SPIE, 2019.

\bibitem{zhang2022adfnet}
Chengkai Zhang, Jichao Jiao, Wei Xu, Ning Li, Min Pang, and Jianye Dong.
\newblock Adfnet: Attention-based fusion network for few-shot rgb-d semantic segmentation.
\newblock In {\em 2022 14th International Conference on Machine Learning and Computing (ICMLC)}, pages 91--96, 2022.

\bibitem{zhang2020sg}
Xiaolin Zhang, Yunchao Wei, Yi Yang, and Thomas~S Huang.
\newblock Sg-one: Similarity guidance network for one-shot semantic segmentation.
\newblock {\em IEEE transactions on cybernetics}, 50(9):3855--3865, 2020.

\bibitem{zhao2017pyramid}
Hengshuang Zhao, Jianping Shi, Xiaojuan Qi, Xiaogang Wang, and Jiaya Jia.
\newblock Pyramid scene parsing network.
\newblock In {\em Proceedings of the IEEE Conference on Computer Vision and Pattern Recognition}, pages 2881--2890, 2017.

\bibitem{zhao2023bmdenet}
Ying Zhao, Kechen Song, Yiming Zhang, and Yunhui Yan.
\newblock Bmdenet: Bi-directional modality difference elimination network for few-shot rgb-t semantic segmentation.
\newblock {\em IEEE Transactions on Circuits and Systems II: Express Briefs}, 2023.

\bibitem{zhu2017unpaired}
Jun-Yan Zhu, Taesung Park, Phillip Isola, and Alexei~A Efros.
\newblock Unpaired image-to-image translation using cycle-consistent adversarial networks.
\newblock In {\em Proceedings of the IEEE international conference on computer vision}, pages 2223--2232, 2017.

\bibitem{zuiderveld1994contrast}
Karel Zuiderveld.
\newblock Contrast limited adaptive histogram equalization.
\newblock {\em Graphics gems}, pages 474--485, 1994.

\end{thebibliography}
}
\end{document}

%%%%%%%%% TITLE 
\title{Supplementary Materials for Generative Model-Based Fusion for Improved Few-Shot Semantic Segmentation of Infrared Images}  
\maketitle

These supplementary materials provide detailed descriptions of the Few-Shot Segmentation (FSS) architectures for both the baseline and the proposed networks. Additionally, they include the ablation studies mentioned in the main text.

\subsection*{A. FSS Baseline Architecture}
\begin{figure*}[!b]
    \centering
    \vspace{3pt}
    \includegraphics[width=.95\textwidth]{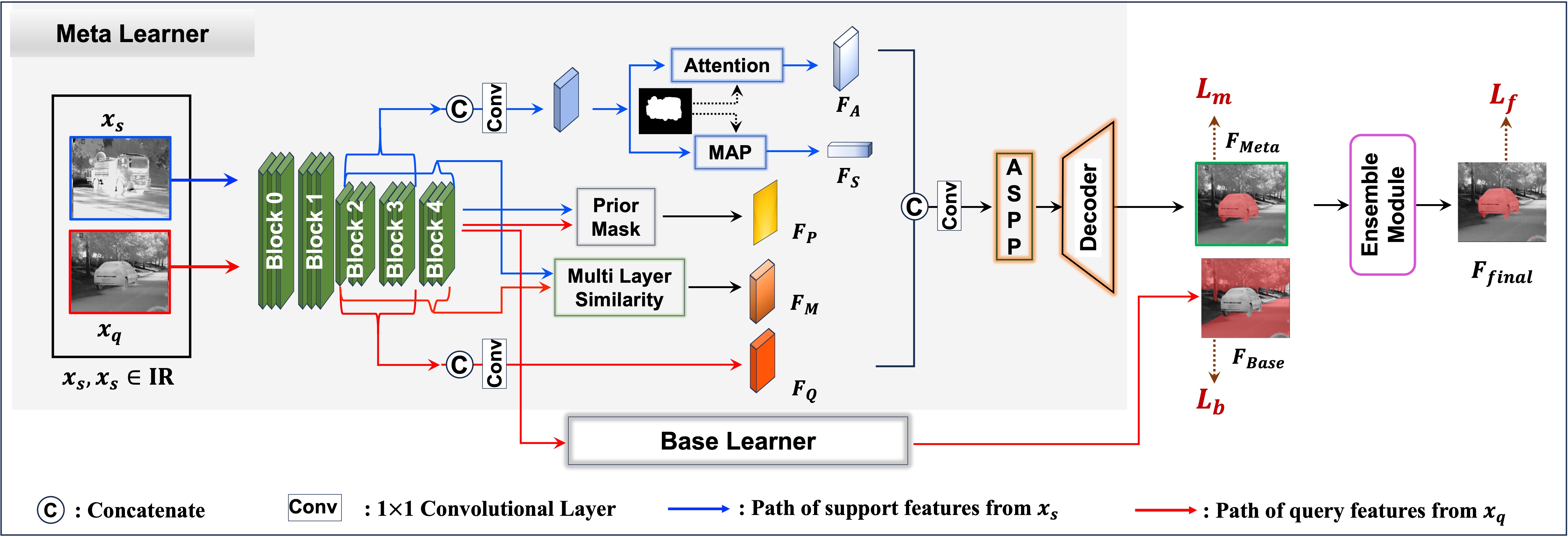}
    \caption{Illustration of the architecture of our baseline (MSANet)~\cite{iqbal2022msanet}.}
    \label{fig:FSS_baseline}
\end{figure*}

The architecture of the baseline is depicted in~\cref{fig:FSS_baseline}, featuring a meta-learner, a base learner, and an ensemble module. The base learner, PSPNet~\cite{zhao2017pyramid}  with a ResNet-50/101~\cite{he2016deep} backbone, is trained in a supervised manner on base classes that are already known, yielding the prediction \( \mathit{F}_{\text{Base}} \) from the query set. Its role is to predict regions of base classes in query images and suppress falsely activated regions of base categories in the meta learner output. The base learner trains the encoder to extract essential features from support and query sets, enabling the meta learner to focus on discerning relationships between these features.

The meta-learner leverages intermediate- and high-level features from the support and query sets to generate five key features to evaluate their relationship. All generated features are concatenated and processed through an atrous spatial pyramid pooling (ASPP) module~\cite{chen2014semantic}, followed by a decoder to produce a binary meta prediction mask \( \mathit{F}_{\text{Meta}} \).  

\begin{figure}[t]
    \centering
    \includegraphics[width=0.45\textwidth]{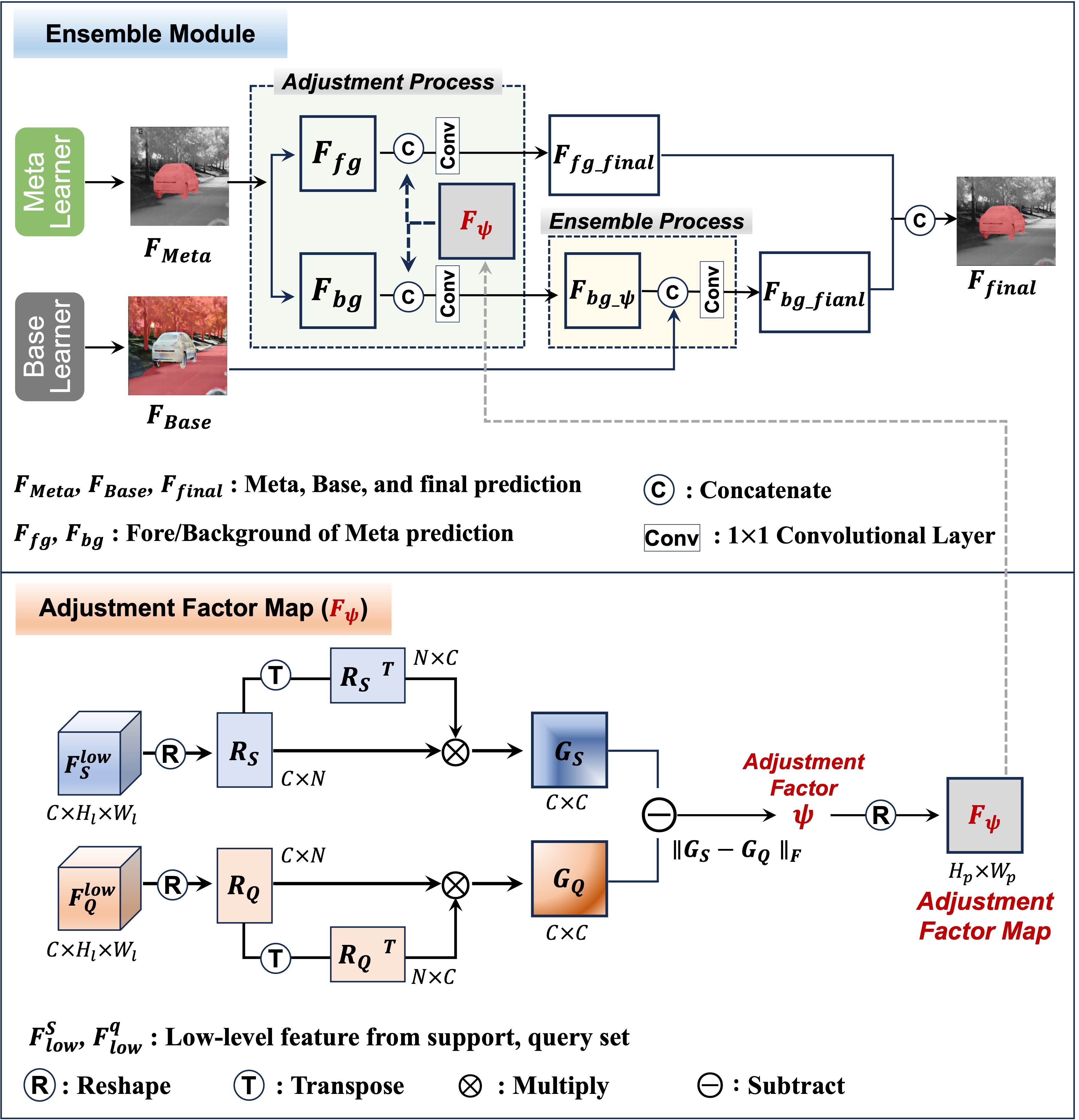}
    \caption{Illustration of a standard ensemble module.}
    \label{fig:ensemble}
\end{figure}

The ensemble module (\cref{fig:ensemble}) integrates these two predictions to produce the final foreground and background probability maps, culminating in the generation of \( \mathit{F}_{\text{final}} \). It initially estimates the scene differences between query-support image pairs by calculating the Gram matrices of the support and query images using low-level features \( \mathit{F}_{\text{S}}^{\mathit{low}}, \mathit{F}_{\text{Q}}^{\mathit{low}} \in \mathbb{R}^{C \times H_l \times W_l} \) extracted from the shared encoder blocks during the training of the meta learner, where \( C \), \( H_l \), and \( W_l \) are the dimensions of the low-level features. The Gram matrices are calculated as follows:

\begin{equation}
G_S = R_S R_S^T \in \mathbb{R}^{C \times C}
\end{equation}
\begin{equation}
G_Q = R_Q R_Q^T \in \mathbb{R}^{C \times C}
\end{equation}
\noindent where \( R_S \) and \( R_Q \) are reshaped tensors of \( \mathit{F}_{\text{S}}^{\mathit{low}} \) and \( \mathit{F}_{\text{Q}}^{\mathit{low}} \), which have dimensions \( C \times N \) (with \( N = H_l \times W_l \)). The Frobenius norm is then computed on the difference between these Gram matrices to derive the adjustment factor map, which guides the adjustment process as calculated:
\begin{equation}
\mathit{F}_{\mathit{\psi}} = Reshape(\|G_S - G_Q\|_{F}) \in \mathbb{R}^{H_p \times W_p}
\end{equation}
\noindent where \(\|\cdot\|_{F}\) indicates the Frobenius norm, and \( Reshape \) is a function reshaping the input tensor to the size of \( H_p \times W_p \), which are the dimensions of the meta and base predictions.

In the adjustment process, the foreground and background of the meta prediction are separately concatenated with the adjustment factor map, followed by a \( 1 \times 1 \) convolutional layer, yielding \( \mathit{F}_{\mathit{fg\_final}} \) and \( \mathit{F}_{\mathit{bg\_\psi}} \). Afterward, in the ensemble process, the base prediction map from the base learner and the adjusted background map \( \mathit{F}_{\mathit{bg\_\psi}} \) are ensembled through concatenation and a convolutional layer, yielding \( \mathit{F}_{\mathit{bg\_final}} \). Finally, the final prediction map \( \mathit{F}_{\mathit{final}} \) is obtained by concatenating \( \mathit{F}_{\mathit{fg\_final}} \) and \( \mathit{F}_{\mathit{bg\_final}} \). 

\subsection*{B. Proposed Overall Architecture with Auxiliary Information}
\begin{figure*}[b]
    \centering
    \vspace{3pt}
    \includegraphics[width=0.97\textwidth]{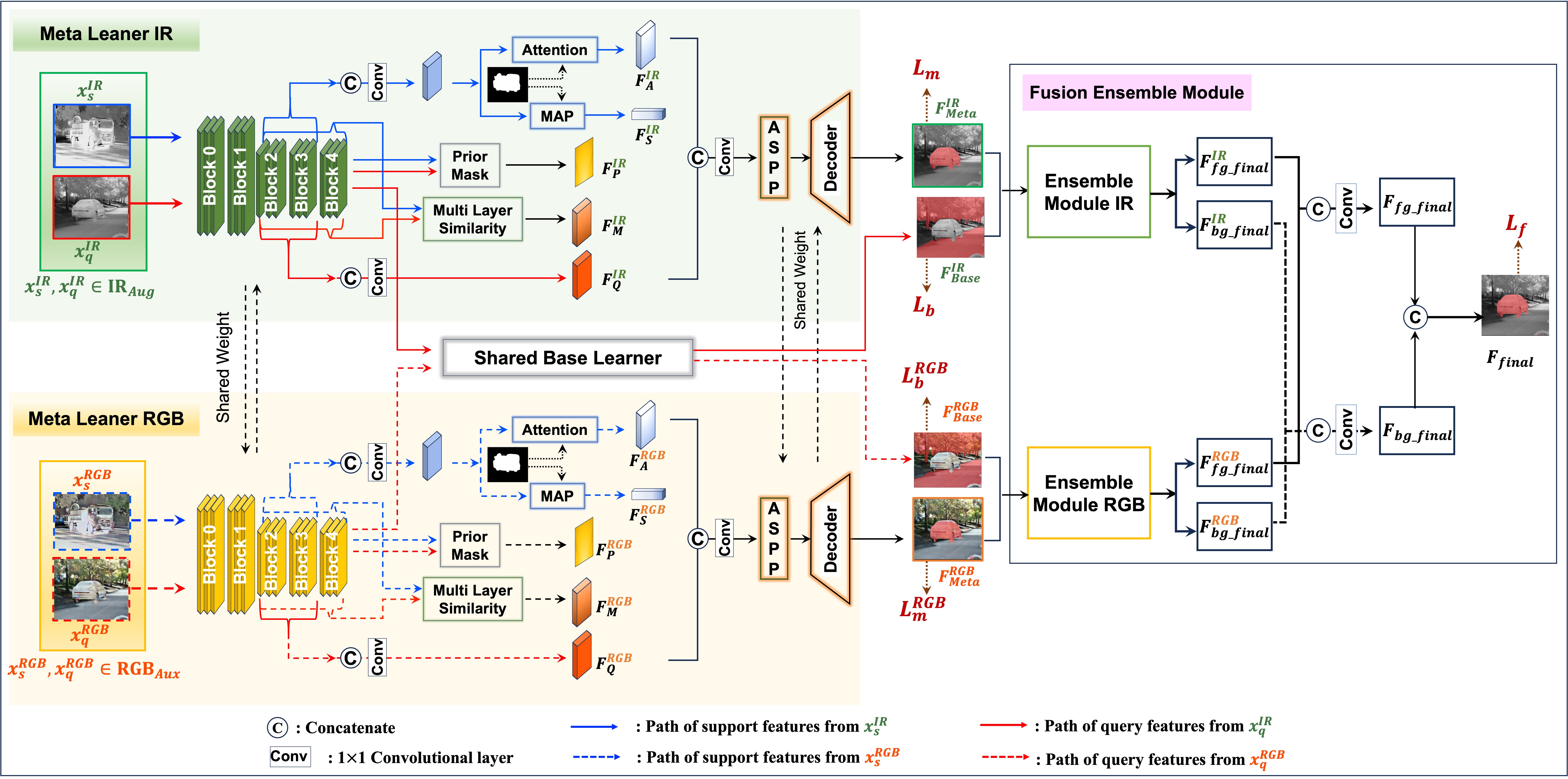}
    \caption{The overall architecture of the proposed methods}
    \label{fig:FSS_proposed_method}
\end{figure*}

In the proposed methods, depicted in~\cref{fig:FSS_proposed_method}, two identical meta-learners, one for each domain, generate meta predictions \( F_{\text{Meta}}^{IR} \) and \( F_{\text{Meta}}^{RGB} \) respectively, utilizing shared encoder, ASPP, and decoder modules. A shared base learner produces predictions \( F_{\text{Base}}^{IR} \) and \( F_{\text{Base}}^{RGB} \). 

The proposed IR-RGB fusion ensemble module comprises two identical ensemble modules for each domain. 
Each ensemble module integrates the meta prediction and base prediction with their own adjustment factor map. The ensembled foreground prediction maps from the IR and RGB domains are then merged with \( 1 \times 1 \) convolutional layers, following the same process for background predictions. The proposed fusion ensemble module complements results from each domain to produce the final foreground/background probability maps and \( F_{\text{final}} \). 

%------------------------------------------------------------------------
 \begin{figure}[b]
  \centering
   \vspace{-.5cm}
   \includegraphics[width=8.0cm, height=7.5cm]{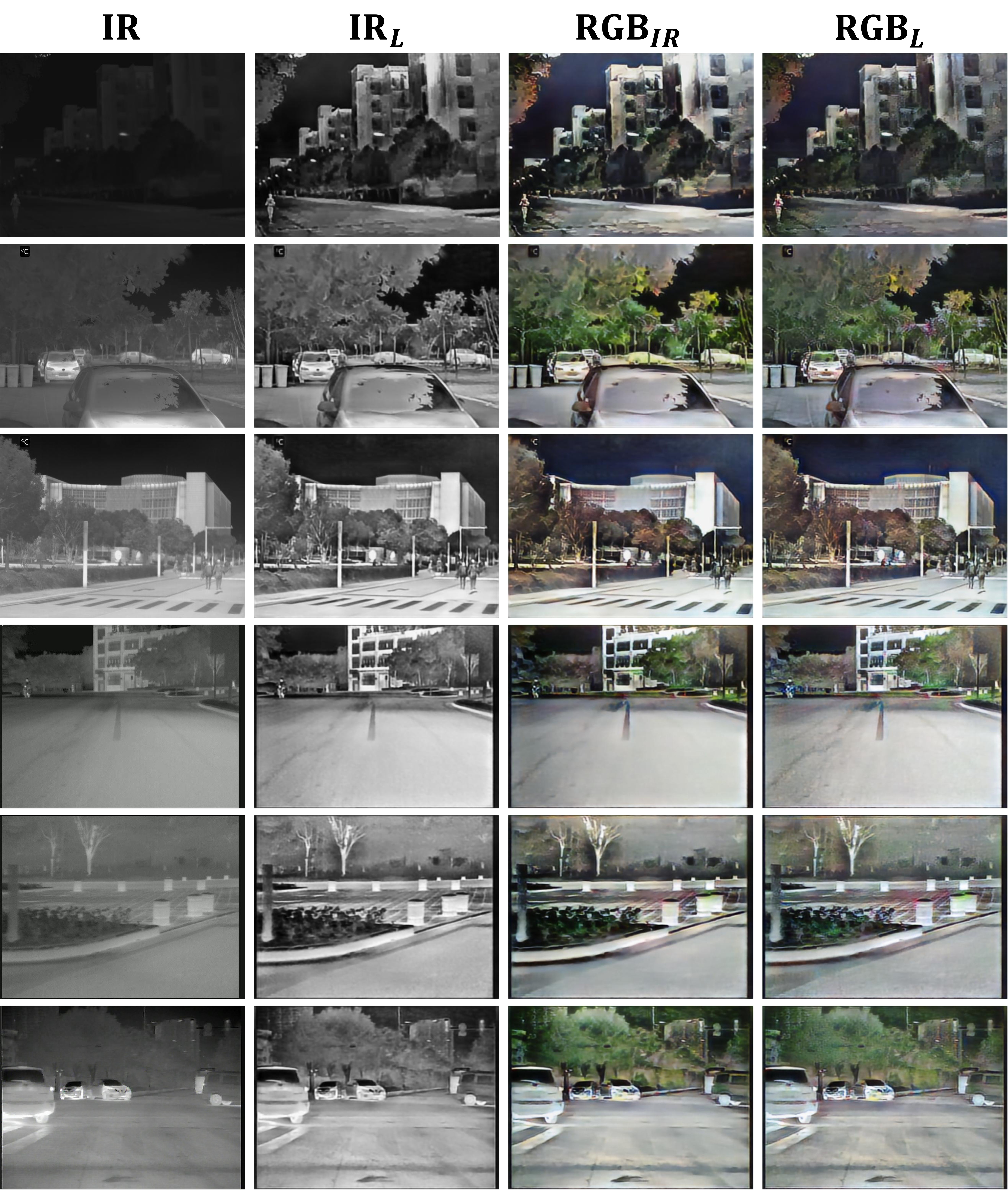}
   \vspace{-0.2cm}
   \caption{The results of I2I translations using SynDiff. The first three rows display samples from the SODA dataset, while the fourth row onwards exhibits samples from the SCUTSEG dataset.}
   \label{fig:qual_syndiff}
\end{figure}
%------------------------------------------------------------------------

%%%% Table of the  with SOTA FSS models %%%%
\begin{table*}[t]
  % \normalsize
  \fontsize{15}{17}\selectfont  
  \centering
  \resizebox{\textwidth}{!}{%
  \renewcommand{\arraystretch}{1.3}
  \begin{tabular}{ c | >{\raggedleft\arraybackslash}p{3.5cm} | c c c c c c | c c c c c c | c c c c c c | c c c c c c}
  
    \toprule
        \multirow{3}{*}{\textbf{Data}} & \multirow{3}{*}{\textbf{Model}} & \multicolumn{12}{c}{\textbf{ResNet-50}} \vline & \multicolumn{12}{c}{\textbf{ResNet-101}} \\ \cline{3-26}
        & & \multicolumn{6}{c}{\textbf{1-shot}} \vline & \multicolumn{6}{c}{\textbf{5-shot}} \vline  & \multicolumn{6}{c}{\textbf{1-shot}} \vline & \multicolumn{6}{c}{\textbf{5-shot}} \\ 
        & & Fold-0 & Fold-1 & Fold-2 & Fold-3 & \textbf{MIoU} & \textbf{FB-IoU} 
        & Fold-0 & Fold-1 & Fold-2 & Fold-3 & \textbf{MIoU} & \textbf{FB-IoU} 
        & Fold-0 & Fold-1 & Fold-2 & Fold-3 & \textbf{MIoU} & \textbf{FB-IoU} 
        & Fold-0 & Fold-1 & Fold-2 & Fold-3 & \textbf{MIoU} & \textbf{FB-IoU}\\
    \midrule
     \multirow{7}{*}{\textbf{SODA~\cite{li2020segmenting}}} 

     % PFENet
     & PFENet~\cite{tian2020prior} &
     32.10 & 24.50 & 32.10 & 32.75 & 30.36 & 55.00 & 
     37.71 & 28.16 & 44.24 & 39.83 & 37.49 & 61.30 & 
     35.16 & 26.78 & 45.38 & 38.71 & 36.51 & 58.72 & 
     41.74 & 33.39 & 59.45 & 49.04 & 45.91 & 66.63  \\ 

     % HSNet
     &  HSNet~\cite{min2021hypercorrelation} &
     33.74 & 23.97 & 33.74 & 35.24 & 31.67 & 58.43 & 
     40.77 & 29.79 & 51.03 & 41.31 & 40.73 & 64.25 & 
     36.06 & 26.07 & 40.50 & 40.12 & 35.69 & 60.69 & 
     43.17 & 33.84 & 55.12 & 47.83 & 44.99 & 66.50  \\ 

     % BAM
     & BAM~\cite{lang2022learning} &
     38.65 & 35.32 & 42.80 & 49.47 & 41.56 & 65.94 & 
     43.89 & 39.54 & 52.42 & \underline{61.73} & 49.40 & 71.77 & 
     39.21 & 36.51 & 51.98 & 53.77 & 45.37 & 68.26 & 
     46.85 & 40.63 & 59.28 & 61.45 & 52.05 & 73.68  \\ 

     % VAT
     & VAT~\cite{hong2022cost} &
     35.40 & 26.65 & 38.88 & 35.12 & 34.01 & 59.58 & 
     40.51 & 31.68 & 48.51 & 43.12 & 40.95 & 63.39 & 
     36.63 & 30.18 & 45.89 & 38.71 & 37.85 & 62.33 & 
     42.97 & 36.27 & 56.50 & 47.86 & 45.90 & 67.20  \\ 

     % MSANet
     & MSANet~\cite{iqbal2022msanet} &
     \underline{43.58} & \underline{37.35} & \underline{47.45} & \underline{55.34} & \underline{45.93} & \underline{70.15} & 
     \underline{48.15} & \underline{42.00} & \underline{61.23} & 59.39 & \underline{52.69} & \underline{74.44} &     
     \underline{42.39} & \textbf{38.73} & \textbf{53.59} & \underline{58.11} & \underline{48.20} & \underline{70.72} & 
     \underline{48.20} & \textbf{42.72} & \underline{64.34} & \underline{63.88} & \underline{54.78} & \underline{75.18} \\   

     % MSI
     & MSI~\cite{moon2023msi} &
     32.32 & 24.98 & 42.19 & 37.82 & 34.33 & 59.22 & 
     37.07 & 29.68 & 51.03 & 45.28 & 40.76 & 62.75 & 
     31.29 & 28.60 & 46.65 & 40.78 & 36.83 & 61.50 & 
     36.91 & 34.17 & 54.09 & 48.58 & 43.44 & 65.57  \\ \cline{2-26}

     % Our method3
     \rowcolor{gray!20!white}
     \cellcolor{white} & Ours (Method3) &
     \textbf{44.01} & \textbf{38.48} & \textbf{50.60} & \textbf{61.19} & \textbf{48.57} & \textbf{71.81} & 
     \textbf{50.92} & \textbf{42.79} & \textbf{60.09} & \textbf{65.36} & \textbf{54.79} & \textbf{75.98} &     
     \textbf{45.38} & \underline{38.04} & \underline{52.51} & \textbf{60.90} & \textbf{49.21} & \textbf{72.21} & 
     \textbf{51.96} & \underline{42.50} & \textbf{65.64} & \textbf{65.74} & \textbf{56.46} & \textbf{76.14}  \\ 
     
     % SCUTSEG Results
     \hline
     \multirow{7}{*}{\textbf{SCUTSEG~\cite{xiong2021mcnet}}} 
     % PFENet
     & PFENet~\cite{tian2020prior} &
     47.15 & 21.76 & 38.44 & 6.74 & 28.52 & 63.52 & 
     48.43 & 25.65 & 40.92 & 7.51 & 30.63 & 65.68 & 
     49.16 & 26.42 & 37.52 & 12.75 & 31.46 & 65.06 & 
     \underline{53.75} & 33.71 & 39.63 & 23.39 & 37.62 & 67.37  \\ 

     % HSNet
     &  HSNet~\cite{min2021hypercorrelation} &
     36.72 & 20.65 & 27.47 & \underline{14.98} & 24.95 & 61.30 & 
     42.24 & 26.88 & 32.60 & \underline{18.01} & 29.93 & 64.62 & 
     38.46 & 20.99 & 28.34 & 8.66 & 24.11 & 61.38 & 
     45.07 & 27.47 & 34.37 & 14.24 & 30.29 & 64.71  \\ 

     % BAM
     & BAM~\cite{lang2022learning} &
     47.48 & 25.49 & 44.73 & 1.96 & 29.92 & 65.93 & 
     50.72 & 30.84 & 45.99 & 5.16 & 33.18 & 67.91 & 
     \underline{50.47} & 29.61 & 40.30 & 9.47 & 32.46 & 65.35 & 
     52.65 & 38.72 & 44.14 & \textbf{27.53} & \underline{40.76} & \underline{69.04}  \\ 

     % VAT
     & VAT~\cite{hong2022cost} &
     37.28 & 22.18 & 34.14 & 10.2 & 26.13 & 61.36 & 
     44.54 & 27.83 & 36.87 & 12.45 & 30.42 & 63.67 & 
     39.11 & 24.72 & 33.04 & \underline{13.55} & 27.61 & 62.55 & 
     44.44 & 30.54 & 38.80 & 18.56 & 33.08 & 65.12  \\ 

     % MSANet
     & MSANet~\cite{iqbal2022msanet} &

     48.35 & \underline{27.29} & \underline{46.51} & 3.97 & \underline{31.53} & \underline{66.68} & 
     \underline{51.69} & \underline{35.46} & \underline{48.53} & 13.20 & \underline{37.22} & \underline{68.68} &      
     50.38 & \underline{30.18} & \underline{44.68} & 11.83 & \underline{34.27} & \underline{66.96} & 
     52.48 & \underline{39.26} & \underline{47.18} & 21.75 & 40.17 & 68.49 \\   
     
     % MSI
     & MSI~\cite{moon2023msi} &
     39.16 & 25.15 & 32.54 & 7.39 & 26.06 & 62.54 & 
     43.35 & 28.27 & 34.44 & 7.22 & 28.32 & 63.93 & 
     40.72 & 25.40 & 28.53 & 7.69 & 25.59 & 62.03 &   
     44.50 & 28.78 & 30.61 & 6.31 & 27.55 & 63.61  \\ \cline{2-26}

     % Our method3
     \rowcolor{gray!20!white}
     \cellcolor{white} & Ours (Method3) &
     \textbf{55.44} & \textbf{34.30} & \textbf{55.09} & \textbf{16.51} & \textbf{40.33} & \textbf{70.48} & 
     \textbf{57.46} & \textbf{41.75} & \textbf{54.29} & \textbf{28.00} & \textbf{45.38} & \textbf{72.43} &
     \textbf{62.36} & \textbf{37.11} & \textbf{52.91} & \textbf{14.45} & \textbf{41.71} & \textbf{71.14} & 
     \textbf{66.49} & \textbf{46.83} & \textbf{56.03} & \underline{25.90} & \textbf{48.81} & \textbf{73.62} \\ 
     
    \bottomrule
    
  \end{tabular}
  }
  \vspace{-.2cm}
  \caption{Comparison with SOTA methods on the SODA and SCUTSEG datasets under 1-shot and 5-shot settings, using ResNet-50 and ResNet-101 backbone networks. Entries in \textbf{bold} indicate the best performance, while those \underline{underlined} denote the second best.}
  
\vspace{-.3cm}
  \label{tab:SOTA}
\end{table*}

\subsection*{C. Qualitative Evaluation of the Adversarial Generative Diffusion Models}

We demonstrate qualitative results of the generated lightness and RGB images on two different IR datasets, showcasing the effectiveness of our approach in generating realistic and visually appealing results.

\noindent \textbf{Generated Lightness Data for Data Augmentation.} The second column in~\cref{fig:qual_syndiff} presents examples of generated lightness domain images \(\textrm{IR}_{L}\). These images exhibit enhanced contrast compared to the original IR datasets, retaining the essential properties and characteristics. \(\textrm{IR}_{L}\) images provide valuable data augmentation, adding diversity and variations to the training data without extra annotations.

\noindent \textbf{Generated RGB Data for Auxiliary Information.} The third and fourth columns in~\cref{fig:qual_syndiff} depict generated RGB images. The \(\textrm{IR}\) and \(\textrm{IR}_{L}\) are converted into \(\textrm{RGB}_{IR}\) and \(\textrm{RGB}_{L}\), which enrich the channel information. While certain categories like trees, skies, and roads translate clearly, others like cars may lack clear color distinction. Despite potential color ambiguities, these images maintain distinct object contours and contain valuable channel information. 

Note that our fusion model is designed to distill information from synthetic RGB images to improve segmentation, even if the synthetic images are not highly realistic as in conventional image-to-image (I2I) translation problems. Thus, the goodness of the synthetic RGB images is not critical to our developments, and as such was not a focus. In addition, the goodness of the synthetic RGB images may be evaluated by using RGB-IR segmentation methods designed for paired data by substituting true RGB with synthesized RGB data. However, this approach was not pursued due to the scarcity of RGB-T datasets with annotations suitable for FSS settings. Although some datasets containing RGB-T pairs with labels do exist (e.g., PST900~\cite{shivakumar2020pst900}), they are not suitable for FSS tasks as they contain only four categories. While Multi-Spectral-4\(^{i}\), derived from the MFNet~\cite{ha2017mfnet} RGB-T dataset, has been utilized for FSS tasks~\cite{bao2021visible, zhao2023bmdenet}, we were unable to leverage it due to the unavailability of publicly released code. Furthermore, since our datasets do not contain true IR-RGB pairs, both evaluations were not possible.

\subsection*{D. Detailed Comparison with SOTA Methods.}
\noindent \textbf{Implementation Details.}
SOTA models were originally designed for RGB datasets and utilize pre-trained backbone networks on ImageNet~\cite{deng2009imagenet}. For a fair comparison, all models were implemented using the same backbone networks, ResNet-50 and ResNet-101, pre-trained on our IR datasets following the same training protocol as the encoder of the base learner~\cite{lang2022learning}. Consistent training augmentation, optimization strategies, and evaluation procedures were applied across all models, with the exception of learning rates and batch sizes, which were optimized for maximal performance on each individual network.

\noindent \textbf{Results and Analysis.}
~\cref{tab:SOTA} presents further results associated with Table 3 of the main text, detailing each fold for SOTA methods and our proposed approach on SODA and SCUTSEG datasets.
Our proposed method with ResNet-101 achieves the highest mIoU and FB-IoU scores across all folds when compared to SOTA models in both datasets. Although our method yields the highest average scores of four folds, certain individual folds (folds 1 and 2 in the 1-shot setting, fold 1 in the 5-shot setting for SODA, and fold 3 in the 5-shot setting for SCUTSEG) exhibit second-place performance.

% Ablation Study 
\subsection*{E. Ablation Study}

To effectively train the encoder, we utilize the proposed datasets \(\textrm{IR}_{Aug}\) and \(\textrm{RGB}_{Aux}\) during the base learner stage as detailed in the main text. In this ablation study, we compare the base learner's predictions from different encoders to assess their impact. ~\cref{tab:ablation} presents the mean mIoU scores of base learner predictions across four folds for the test set. For both ResNet-50 and ResNet-101 as encoders, the results indicate enhanced mIoU scores with augmented and auxiliary data in both datasets, with all proposed methods. This improvement signifies improved capturing of features from both the support and query sets.

\begin{table}[h]
  \scriptsize
  \centering
  \setlength{\tabcolsep}{5pt}
  \renewcommand{\arraystretch}{0.9}
  \begin{tabular}{c|c|c|c|c}  
\toprule
    & \multicolumn{2}{c}{\textbf{SODA}} \vline  & \multicolumn{2}{c}{\textbf{SCUTSEG}} \\
    \textbf{Method} & \textbf{ResNet-50} &  \textbf{ResNet-101} &  \textbf{ResNet-50} &  \textbf{ResNet-101} \\  
       
\midrule
  Baseline & 54.09             & 56.20             & 43.26             & 45.01 \\
  Method 1 & \underline{55.67} & \textbf{57.32}    & \textbf{47.65}    & \underline{48.41} \\
  Method 2 & 55.66             & 57.00             & 46.79             & 48.39 \\
  Method 3 & \textbf{56.10}    & \underline{57.18} & \underline{47.35} & \textbf{50.81} \\
\bottomrule
    
  \end{tabular}
  \caption{The mean mIoU of the base learner's predictions across four folds for the test set.}
  \label{tab:ablation}
\end{table}

% \clearpage
%%%%%%%%% REFERENCES
{\small
\bibliographystyle{ieee_fullname}

}